%% file: main.tex
\documentclass[11pt]{article}

\usepackage[final]{acl}

\usepackage{times}
\usepackage{latexsym}

\usepackage[T1]{fontenc}

\usepackage[utf8]{inputenc}

\usepackage{microtype}

\usepackage{inconsolata}

\usepackage{graphicx}
\usepackage{enumitem}
\usepackage{booktabs}
\usepackage{todonotes}
\usepackage[most]{tcolorbox}
\usepackage{colortbl}

\definecolor{myblue}{HTML}{4C78A8}
\definecolor{myorange}{HTML}{B6992D}
\definecolor{mygreen}{HTML}{59A14F}
\definecolor{mypurple}{HTML}{8e6bbe}

\newtcolorbox{mytextbox}{
    enhanced,
    colback=gray!5,
    colframe=myblue,
    arc=0.5mm,
    outer arc=0.5mm,
    top=0.5mm, bottom=0.5mm, left=0.5mm, right=0.5mm,
    boxrule=0mm, leftrule=0.5mm,
fontupper=\tiny\raggedright
}
\usepackage{cleveref}
\newcommand{\ours}{\textsc{Chartographer}}

\title{\ours: Counterfactual Chart Generation for \\Evaluating Vision-Language Models}

\author{
  \textbf{Yifan Jiang\textsuperscript{1,2}} \quad
  \textbf{Dae Yon Hwang\textsuperscript{3}} \quad
  \textbf{Jesse C. Cresswell\textsuperscript{3}} \quad
  \textbf{Freda Shi\textsuperscript{1,2}}\\
  \textsuperscript{1}University of Waterloo \quad
  \textsuperscript{2}Vector Institute \quad
  \textsuperscript{3}Layer 6 AI\\
  \small{\texttt{\{yifan.jiang, fhs\}@uwaterloo.ca}} \quad
  \small{\texttt{\{daeyon, jesse\}@layer6.ai}}
}

\begin{document}
\maketitle
\input{sections/00_abstract}

\input{sections/01_introduction}
\input{sections/02_related_work}
\input{sections/03_framework}
\input{sections/04_experiment_setup}
\input{sections/05_results}
\input{sections/06_conclusion}
\input{sections/07_limitations}

\section*{Acknowledgment}
This work is supported in part by the NSERC Discovery Grant RGPIN-2024-04395 and a Canada CIFAR AI Chair Award to FS, and by Mitacs through the Mitacs Accelerate program.
AI assistants were used for coding and proofreading. All AI-assisted outputs were reviewed and verified by the authors.

\bibliography{bib}

\clearpage

\appendix
\input{sections/A-appendix}

\end{document}

%% file: sections/00_abstract.tex
\begin{abstract}
Chart question-answering (QA) benchmarks aim to pose questions that require visual reasoning to correctly answer, but vision-language models (VLMs) can often reach solutions through shortcuts or prior familiarity with a chart or question. To strictly evaluate visual reasoning, we propose \emph{counterfactual charts} where the chart-question task remains fixed, but the underlying data and the corresponding answer are varied. We introduce \textsc{Chartographer}, a framework to reverse engineer charts into executable code, validate reconstruction fidelity, generate counterfactual variants, and derive new answers from executable QA logic. We apply this framework to existing chart QA datasets and evaluate proprietary and open-source VLMs, measuring variant sensitivity and generalizability. Counterfactual charts reveal failures hidden by single-chart performance: VLMs often fail to generalize after answering the original chart correctly. We find that failures are most prevalent when updated charts require novel visual reasoning pathways.
\end{abstract}

%% file: sections/01_introduction.tex
\section{Introduction}

Charts are a compact language for communicating quantitative evidence.
Answering questions based on information in charts requires more than recognizing visual elements: a model must read labels, identify encodings, make comparisons, aggregate values, track trends, and sometimes combine visual evidence with domain conventions \cite{kafle2018dvqa, kahou2017figureqa}.
As vision-language models (VLMs) are increasingly used to understand and summarize scientific papers, financial reports, dashboards, and other data-rich documents, chart question-answering (QA) has become an important test of multimodal reasoning.

Chart QA benchmarks now cover a wide range of synthetic, web-sourced, and scientific visualizations \cite{methani2020plotqa, masry2022chartqabenchmarkquestionanswering, wang2024charxivchartinggapsrealistic}.
However, they largely evaluate fixed chart-question-answer triples where each question is tied to one specific chart and answer. 
Benchmarks aim to measure visual reasoning ability, but often the correct answer can be reached through shortcuts, by exploiting regularities in the question, or relying on parametric knowledge acquired from charts or other sources during pre-training. 
What remains unclear from existing benchmarks is whether a VLM's behavior on the same chart-question task generalizes when the underlying data changes.

In this paper, we introduce \ours, a framework for generating \emph{counterfactual charts} whose answers are recomputed after controlled data changes, to better distinguish visual reasoning ability from potential reliance on shortcuts.
The desired behavior is simple: when the visual evidence changes, the answer should change accordingly.
Our intervention changes answer-relevant data while preserving the chart schema and question task, thereby isolating evidence updating from changes in chart type, layout, visual encoding, or question formulation.
We propose the chart-question \emph{family} as a primary evaluation unit that includes an original chart, its base reconstruction, and counterfactual variants.  
This shifts evaluation from recovering one fixed answer to measuring whether model predictions remain grounded in changed visual evidence.
Each family is validated through a base reconstruction that checks whether the task survives chart-to-code reverse engineering before introducing data changes.
We then compare behavior within the family to determine whether success on the original chart generalizes to counterfactual variants. Our contributions are:

\begin{figure*}[t]
\centering
\includegraphics[width=\textwidth]{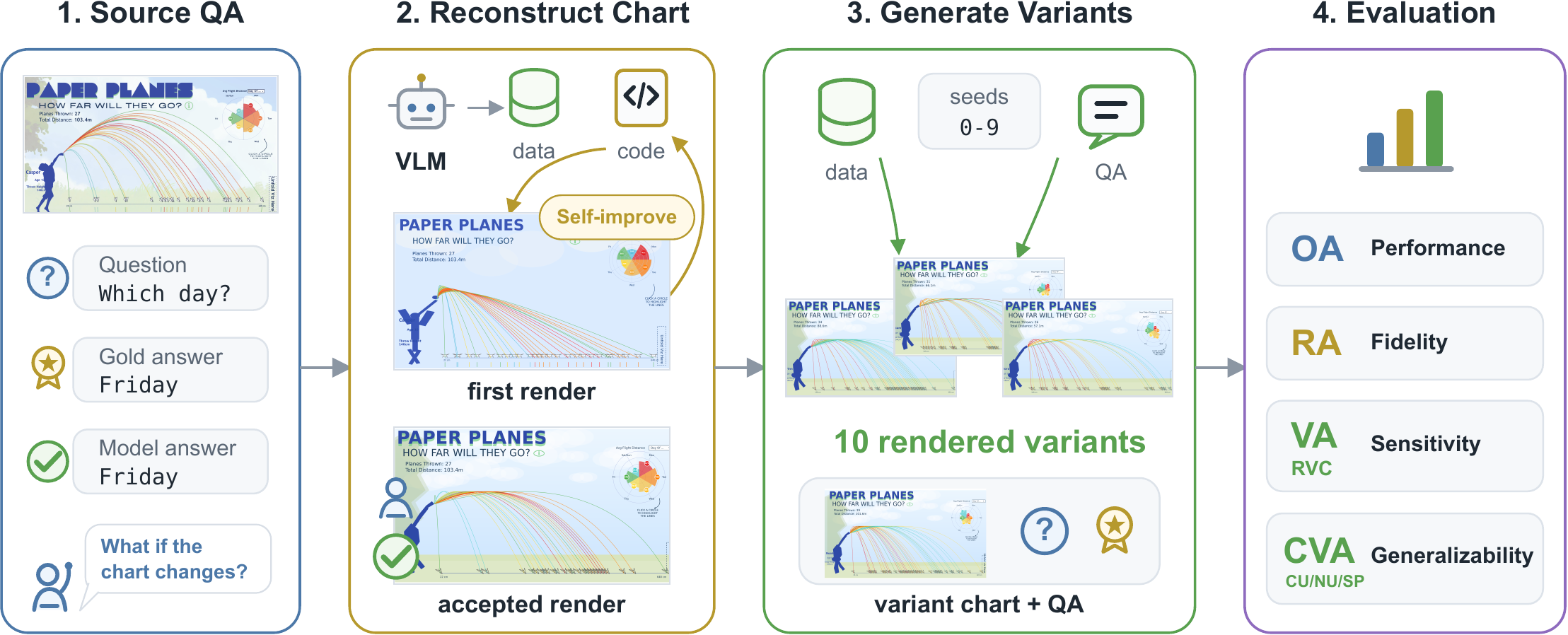}
\caption{The \ours\ pipeline for constructing counterfactual chart-question families. Starting from a source chart QA example, the pipeline reconstructs semantic chart data and chart-rendering code, iteratively revises the render until accepted, generates seed-controlled counterfactual variants, and recomputes answers with executable QA logic. The resulting families test whether success on an original chart-question task generalizes with altered visual evidence. See \Cref{sec:counterfactual-evaluation} for evaluation metric definitions.}
\label{fig:pipeline-overview}
\vspace{-10pt}
\end{figure*}

\begin{itemize}[noitemsep,topsep=0pt,leftmargin=*]
  \item We create a chart-to-code pipeline that converts an individual chart QA example into a counterfactual chart-question family, with iterative reconstruction, human-in-the-loop validation, and executable QA generation.
  \item We propose a counterfactual VLM evaluation framework that covers single-chart performance, reconstruction fidelity, variant sensitivity, and counterfactual generalizability.
  \item We apply this framework to existing chart QA benchmarks, showing that success on original charts often fails to generalize when visual evidence changes.
\end{itemize}

%% file: sections/02_related_work.tex
\vspace{-4pt}
\section{Related Work}
\vspace{-4pt}

\noindent\textbf{Chart QA benchmarks.} 
VLM reasoning ability has been studied through both synthetic and human-generated chart QA benchmarks.
DVQA~\cite{kafle2018dvqa} and FigureQA~\cite{kahou2017figureqa} use controlled chart or figure generation to isolate basic visual reasoning operations, while PlotQA~\cite{methani2020plotqa} focuses on scientific plots.
ChartQA~\cite{masry2022chartqabenchmarkquestionanswering} combines web-sourced charts with human-authored questions requiring more complex visual and logical reasoning.
More recent datasets such as CharXiv~\cite{wang2024charxivchartinggapsrealistic} and ChartMuseum~\cite{tang2026chartmuseumtestingvisualreasoning} move toward realistic scientific figures, human-designed visualizations, and questions that require visual, textual, and multimodal understanding.
This progression has made chart QA more realistic and challenging, but the dominant evaluation unit remains a fixed chart-question-answer triple.
Our work uses ChartQA, CharXiv, and ChartMuseum as source benchmarks, but instead of evaluating only chart-question-answer triples, we convert examples into counterfactual chart-question families to isolate true visual reasoning.

\vspace{2pt}\noindent\textbf{Chart understanding with VLMs.}
VLMs combine visual encoders with large language models to align visual inputs with text~\cite{alayrac2022flamingovisuallanguagemodel}, and recent frontier systems report broad multimodal capabilities~\cite{openai2025gpt5,anthropic2025claude4,google2025gemini25models,bai2025qwen3}.
Charts remain challenging because models must parse visual encodings, recover quantities, follow labels and legends, and perform numerical operations over the extracted evidence.
Prior work aims to improve chart reasoning through intermediate representations, such as screenshot-to-HTML parsing~\cite{lee2023pix2struct}, chart-to-table derendering~\cite{liu-etal-2023-deplot, liu2023matcha}, grounding and reflection~\cite{Xu_2025_ICCV}, and distillation of chart reasoning trajectories~\cite{he-etal-2025-distill}.
Our focus is complementary: we use counterfactual chart-question families to test whether VLM chart reasoning generalizes when the underlying chart data changes. 

\noindent\textbf{Shortcuts and counterfactual evaluation.} 
Shortcut behavior is well documented in language benchmarks: \citet{gururangan-etal-2018-annotation} shows that annotation artifacts in natural language inference data can make labels partially predictable from the hypothesis alone, without access to the premise.
The same concern extends to multimodal evaluation, where visual reasoning benchmarks can reward shortcuts rather than application of visual logic, including reliance on parametric knowledge and superficial linguistic regularities~\cite{hou2025do,chi2025chimeradiagnosingshortcutlearning,xia2025visionary}.
This risk is relevant for charts, where previously encountered images, recognizable templates, and predictable answer distributions may obscure whether a VLM genuinely reasons over visual evidence.
Chart-HQA~\cite{chen2025charthqabenchmarkhypotheticalquestion} introduces hypothetical assumptions over chart questions to probe counterfactual reasoning over chart content. 
CharXiv evaluates charts of similar visual complexity with newly annotated questions as an alternative to templated questions~\cite{wang2024charxivchartinggapsrealistic}.
Other benchmarks vary chart presentation while holding the data fixed: ChartInsights~\cite{wu2024chartinsights} changes color schemes and adds image noise, while ChartR~\cite{chen2026chartr} evaluates blur, noise, watermarks, and annotation removal.
Our work regenerates the charts themselves with altered data, rather than asking textual hypotheticals over a single chart or replacing charts with visually similar ones, thereby requiring models to reason over changed visual evidence.

%% file: sections/03_framework.tex
\vspace{-2pt}
\section{\textsc{Chartographer}: Counterfactual Chart Framework}
\label{sec:framework}
\vspace{-2pt}

Given a chart QA example with a chart image, a question, and an answer, our framework builds a counterfactual chart-question family for the same task, as visualized in \Cref{fig:pipeline-overview}. 
The family contains the original chart, a base reconstruction rendered via reverse-engineered plotting code, and counterfactual variants whose answers are recomputed from the underlying data.
The base reconstruction verifies that the original task survives re-rendering, while the variants test whether predictions remain grounded after alterations to the visual evidence.

\vspace{-2pt}
\subsection{Chart Reconstruction}
\vspace{-2pt}

\vspace{2pt}\noindent\textbf{Chart-to-code reconstruction.}
The first step reverse engineers a chart image into executable plotting code, drawing on the broader idea of recovering structured data or rendering code from visual inputs~\cite{liu2023matcha,yang2025chartmimic}.
For each source chart, a VLM produces semantic chart data and chart-rendering code: the semantic data records quantities, labels, categories, and groups as seen in the original chart, while the code reproduces visual encodings, layout choices, and rendering logic.
This separation is what makes counterfactual generation possible: data values can vary while the chart theme and purpose remain unchanged.
Appendix~\ref{sec:appendix-reconstruction} provides additional reconstruction details.

\vspace{2pt}\noindent\textbf{Self-improvement.}
Reverse engineering complex charts into data assets and code for re-rendering rarely works perfectly on the first attempt.
Hence, we iteratively improve the conversion through a VLM-led self-improvement loop.
After the initial data/code pair is generated, the chart is re-rendered and then compared with the original. 
The VLM diagnoses differences and flaws in the reconstruction, creates a plan for improvement, and then executes that plan to generate another version of data and code. 
This loop is continued until the VLM raises no significant concerns, or to a maximum number of iterations (we cap at five).
Appendix~\ref{sec:appendix-self-improvement-human-validation} provides additional details on self-improvement.

\vspace{2pt}\noindent\textbf{Human-in-the-loop validation.}
After self-improvement, the VLM classifies its work as acceptable, unsuccessful, or in need of human feedback. 
Some reconstructions require human judgment because labels are ambiguous, values are unreadable, or the intended visual encoding is not recoverable from pixels alone.
Human reviewers inspect difficult but promising cases, approve usable reconstructions, reject low-fidelity outputs, and record assumptions for downstream generation. 
Appendix~\ref{sec:appendix-self-improvement-human-validation} provides additional details on human validation.

\vspace{-2pt}
\subsection{Counterfactual Chart Generation}
\vspace{-2pt}
For each accepted reconstruction, we create a seed-controlled data generator and use it to generate ten counterfactual charts that resemble the original but reflect the altered data.
The generator preserves the chart schema, rendering constraints, and domain assumptions while changing the data in meaningful ways.
Appendix~\ref{sec:appendix-counterfactual-data-generation} provides additional details on counterfactual data and chart generation.

\vspace{-2pt}
\subsection{QA Generation}
\vspace{-2pt}

Counterfactual charts require valid questions and answers that reflect their altered data.
For each accepted reconstruction, we create executable QA logic that computes answers directly from the underlying data, not the visual chart. 
The original question is kept whenever it remains valid, and is rewritten only when necessary to maintain coherence.
This makes large-scale counterfactual labeling feasible without manually annotating every generated chart.
Appendix~\ref{sec:appendix-qa-generation} provides additional details on QA generation.

\vspace{-2pt}
\subsection{Counterfactual Family Evaluation}
\label{sec:counterfactual-evaluation}
\vspace{-2pt}

Each family contains three components: the original chart, the base reconstruction, and counterfactual variants.
The base reconstruction is a control: it tests whether the chart-question task survives re-rendering with unaltered data.
The counterfactual variants are the intervention: they test whether a VLM remains grounded when visual evidence changes for the same chart-question task.
We therefore separate single-chart performance, reconstruction fidelity, variant sensitivity, and counterfactual generalizability via five metrics:
\begin{itemize}[noitemsep,topsep=0pt,leftmargin=*]
  \item \textbf{Original accuracy (OA)}: Accuracy on the original charts.
  \item \textbf{Reconstruction accuracy (RA)}: Accuracy on base reconstructions with unaltered data.
  \item \textbf{Variant accuracy (VA)}: Average accuracy on counterfactual variants with changed visual evidence and updated QA pairs.
  \item \textbf{Relative variant change (RVC)}: Computed as $100\cdot(\mathrm{VA}-\mathrm{OA})/\mathrm{OA}$, $\mathrm{RVC}$ normalizes sensitivity to variant charts.
  \item \textbf{Conditional variant accuracy (CVA)}: Counterfactual generalizability measured as VA restricted to families where the model answered the original chart correctly.
\end{itemize}
Appendix~\ref{sec:appendix-metric-definitions} defines the metrics and aggregation procedure formally.

To diagnose failed generalization, we analyze variants where the model answered the original chart correctly.
Within this conditional subset, each prediction falls into one of three update outcomes. A \textbf{correct update (CU)} answers the variant correctly. A \textbf{stale prediction (SP)} repeats the original prediction and is wrong. A \textbf{noisy update (NU)} changes the answer but remains incorrect.
We pair these metrics with failure case studies by inspecting charts, variants, labels, generated rationales, and model predictions from persistently difficult families.

%% file: sections/04_experiment_setup.tex
\begin{table*}[t!]
\centering
\footnotesize
\setlength{\tabcolsep}{3.8pt}
\resizebox{\textwidth}{!}{
\begin{tabular}{lrrrrrrrrrrrr}
\toprule
 & \multicolumn{4}{c}{ChartQA} & \multicolumn{4}{c}{CharXiv} & \multicolumn{4}{c}{ChartMuseum} \\
\cmidrule(lr){2-5} \cmidrule(lr){6-9} \cmidrule(lr){10-13}
Model & OA & RA & VA & RVC & OA & RA & VA & RVC & OA & RA & VA & RVC \\
\midrule
\multicolumn{13}{l}{\textit{Proprietary models}} \\
Claude Haiku 4.5 & 0.907 & 0.933 & 0.878 & -3.2 & 0.541 & 0.608 & 0.574 & +6.2 & 0.472 & 0.451 & 0.357 & $^{**}$-24.3 \\
Claude Sonnet 4.6 & 0.927 & 0.933 & 0.905 & -2.3 & 0.662 & 0.750 & 0.697 & +5.2 & 0.690 & 0.704 & 0.541 & $^{***}$-21.6 \\
Gemini 2.5 Flash & 0.933 & 0.933 & 0.891 & $^{**}$-4.5 & 0.608 & 0.669 & 0.647 & +6.4 & 0.641 & 0.606 & 0.493 & $^{***}$-23.1 \\
Gemini 2.5 Pro & 0.947 & 0.927 & 0.897 & $^{**}$-5.2 & 0.682 & 0.750 & 0.726 & +6.3 & 0.732 & 0.704 & 0.529 & $^{***}$-27.8 \\
GPT-4.1 & 0.927 & 0.900 & 0.869 & $^{**}$-6.3 & 0.547 & 0.574 & 0.534 & -2.3 & 0.507 & 0.549 & 0.392 & $^{**}$-22.6 \\
GPT-4.1 mini & 0.880 & 0.900 & 0.861 & -2.1 & 0.412 & 0.622 & 0.476 & +15.6 & 0.528 & 0.528 & 0.446 & $^{*}$-15.5 \\
GPT-4o & 0.860 & 0.887 & 0.841 & -2.2 & 0.432 & 0.426 & 0.389 & -10.2 & 0.345 & 0.366 & 0.287 & -16.7 \\
GPT-5.4 & 0.927 & 0.947 & 0.916 & -1.2 & 0.770 & 0.757 & 0.803 & +4.3 & 0.704 & 0.718 & 0.554 & $^{***}$-21.3 \\
GPT-5.4 mini & 0.940 & 0.940 & 0.917 & -2.5 & 0.676 & 0.696 & 0.696 & +3.0 & 0.486 & 0.514 & 0.425 & -12.5 \\
\textbf{Proprietary average} & \textbf{0.916} & \textbf{0.922} & \textbf{0.886} & \textbf{-3.3} & \textbf{0.592} & \textbf{0.650} & \textbf{0.616} & \textbf{+3.8} & \textbf{0.567} & \textbf{0.571} & \textbf{0.447} & \textbf{-20.6} \\
\midrule
\multicolumn{13}{l}{\textit{Open-source models}} \\
Gemma 4 E4B IT & 0.860 & 0.853 & 0.839 & -2.5 & 0.365 & 0.439 & 0.411 & +12.6 & 0.289 & 0.261 & 0.251 & -12.9 \\
InternVL3-8B & 0.847 & 0.853 & 0.816 & -3.6 & 0.311 & 0.318 & 0.304 & -2.2 & 0.268 & 0.239 & 0.223 & -16.6 \\
LLaVA-OneVision-Qwen2-7B & 0.707 & 0.680 & 0.612 & $^{**}$-13.4 & 0.250 & 0.189 & 0.196 & -21.6 & 0.190 & 0.211 & 0.164 & -13.7 \\
Pixtral-12B & 0.827 & 0.787 & 0.767 & $^{*}$-7.3 & 0.372 & 0.365 & 0.317 & -14.7 & 0.246 & 0.254 & 0.211 & -14.6 \\
Qwen2.5-VL-7B-Instruct & 0.840 & 0.880 & 0.790 & -6.0 & 0.338 & 0.372 & 0.343 & +1.6 & 0.275 & 0.211 & 0.223 & -18.7 \\
Qwen3-VL-8B-Instruct & 0.907 & 0.907 & 0.882 & -2.7 & 0.480 & 0.568 & 0.506 & +5.5 & 0.380 & 0.423 & 0.339 & -10.7 \\
\textbf{Open-source average} & \textbf{0.831} & \textbf{0.827} & \textbf{0.784} & \textbf{-5.9} & \textbf{0.352} & \textbf{0.375} & \textbf{0.346} & \textbf{-3.1} & \textbf{0.275} & \textbf{0.267} & \textbf{0.235} & \textbf{-14.5} \\
\bottomrule
\end{tabular}
}
\caption{Main accuracy metrics for each model and dataset. Metrics, defined in \Cref{sec:counterfactual-evaluation}, summarize base performance, reconstruction fidelity, and variant sensitivity. Average rows report arithmetic means within each model group. Significance stars on RVC mark per-model two-sided paired sign-flip permutation tests for original-to-variant accuracy change over chart-question families: $^{*}p<.05$, $^{**}p<.01$, $^{***}p<.001$.}
\label{tab:main-results}
\end{table*}

\vspace{-4pt}
\section{Experiment Setup}
\vspace{-4pt}

\noindent\textbf{Datasets.} We use evaluation splits from three chart QA sources: the ChartQA validation split~\cite{masry2022chartqabenchmarkquestionanswering}, the CharXiv validation split~\cite{wang2024charxivchartinggapsrealistic}, and the ChartMuseum development split~\cite{tang2026chartmuseumtestingvisualreasoning}.
These sources provide a varied testbed for chart-question tasks across conventional charts, scientific figures, and human-designed visualizations.
We randomly sample 462 chart QA tasks from these splits and exclude charts that are unsuitable for chart-to-code reconstruction, such as cases with ambiguous labels, occluded values, or visual encodings that do not support controlled data edits.
After filtering, the source sample contains 440 chart QA tasks across 321 of 335 unique charts (95.8\%).
The excluded charts primarily have extremely dense values or custom encodings that our plotting stack cannot reproduce reliably. The retained set still includes diverse sources, such as maps, timelines, scatter plots, and infographics.
Each retained task is converted into a counterfactual chart-question family with the model-assisted pipeline described in Section~\ref{sec:framework}.
Appendix~\ref{sec:appendix-source-data} reports the per-dataset filtered counts.

\noindent\textbf{Models.} We evaluate instruction-tuned VLMs from both proprietary and open-source model families.
The proprietary set includes Claude models~\cite{anthropic2025haiku45,anthropic2026sonnet46}, Gemini models~\cite{google2025gemini25models}, and GPT-family models~\cite{openai2024gpt4o,openai2025gpt41,openai2026gpt54,openai2026gpt54mini} accessed through their respective APIs.
The open-source set is run locally and includes Gemma~\cite{google2026gemma4}, InternVL~\cite{zhu2025internvl3}, LLaVA~\cite{li2024llavaonevision}, Pixtral~\cite{agrawal2024pixtral12b}, and Qwen variants~\cite{bai2025qwen3,bai2025qwen25vl}.
Appendix~\ref{sec:appendix-models} lists the model set and inference settings.

\noindent\textbf{Prompting.} All models receive a chart image and a question, with the prompt template fixed across all models and tasks.
The accuracy metrics and update outcomes use only the extracted final answer. 
We analyze rationales separately in the failure diagnostics of Section~\ref{sec:update-failure-analysis}.
Appendix~\ref{sec:appendix-prompting-evaluation} describes the prompt format and answer extraction.

\noindent\textbf{Evaluation.} For original charts, predictions are evaluated against the benchmark labels.
For base reconstructions and counterfactual variants, predictions are evaluated against labels produced by the executable QA logic.
We use binary accuracy as the base correctness metric, employing an LLM judge to determine whether the prediction is equivalent to the target answer.
We test robustness to judge choice by rescoring all 25,560 ChartMuseum decisions with two additional judges. Pairwise agreement among the three judges is 98.05--98.21\%, and the resulting aggregate metrics are nearly identical.
We report the metrics defined in Section~\ref{sec:counterfactual-evaluation} and use per-model two-sided paired sign-flip permutation tests over chart-question families for OA-to-VA comparisons.
Appendix~\ref{sec:appendix-prompting-evaluation} provides additional details on the evaluation.

%% file: sections/05_results.tex
\begin{figure*}[t]
    \centering
    \begin{minipage}[c]{0.49\textwidth}
        \centering
        \begin{minipage}[c]{0.65\textwidth}
            \centering
            \includegraphics[width=\textwidth]{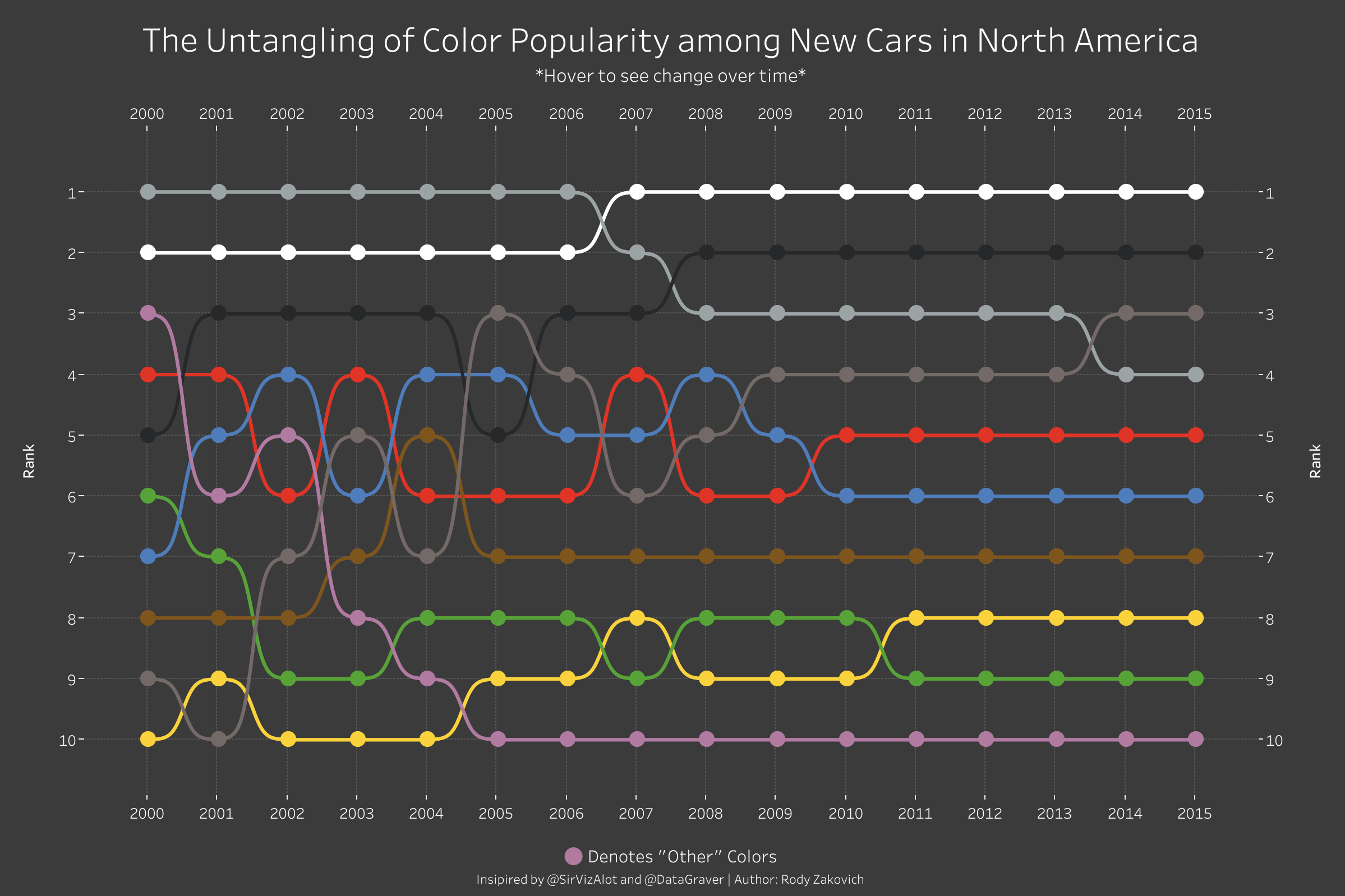}
        \end{minipage}
        \hfill
        \begin{minipage}[c]{0.33\textwidth}
            \begin{mytextbox}
                \textbf{Question:} How many colors were always ranked above red? \ \textbf{Gold:} 2\\
                \textbf{GPT-5.4 mini:} Red is the line that starts at rank 4 in 2000, moves between ranks 4--6 over the years, and is \textcolor{red}{never above} the top-ranked colors like white, \textcolor{red}{black, gray}, or silver when compared year by year. Counting the colors that are always ranked above red throughout the chart gives 4. \textbf{Answer: 4}
            \end{mytextbox}
        \end{minipage}
    \end{minipage}
    \hfill
    \vrule{} 
    \hfill
    \begin{minipage}[c]{0.49\textwidth}
        \centering
        \begin{minipage}[c]{0.65\textwidth}
            \centering
            \includegraphics[width=\textwidth]{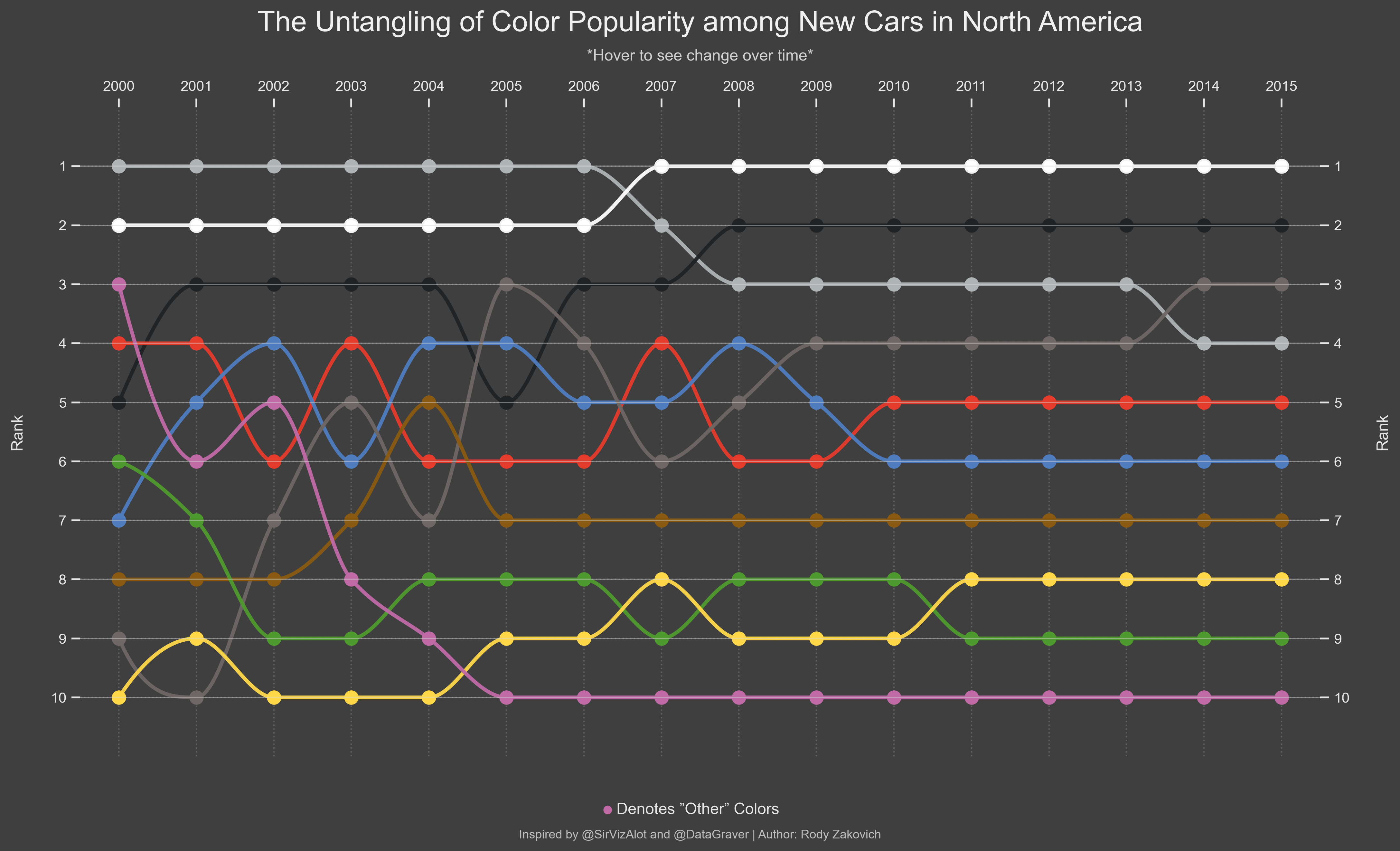}
        \end{minipage}
        \hfill
        \begin{minipage}[c]{0.33\textwidth}
            \begin{mytextbox}
                \textbf{Question:} How many colors were always ranked above red? \ \textbf{Gold:} 2\\
                \textbf{GPT-5.4 mini:} ``Always ranked above red'' means the colors that had a better rank than red in every year shown. Red moves between ranks 4--6 across the years, so only the colors that stay above red in all years count. Those colors are white and silver/gray. \textbf{Answer: 2}
            \end{mytextbox}
        \end{minipage}
    \end{minipage}
    
    \caption{Example where GPT-5.4 mini solves the reconstruction (right), but not the original (left). The reconstruction changes some details such as font sizes, curve shapes, and proportions, and has higher resolution because it was generated by code, not scraped from public sources. The flawed observation is shown in \textcolor{red}{red text}.}
    \label{fig:reconstruction-1}
    \vspace{-10pt}
\end{figure*}

\vspace{-4pt}
\section{Results}
\label{sec:results}
\vspace{-4pt}

Counterfactual chart-question families expose a failure mode that single-chart QA cannot measure: a model may answer the original chart's question correctly but fail to generalize when visual evidence changes.
To examine this failure mode, we first account for reconstruction artifacts by checking that the base reconstructions preserve the original benchmark tasks (RA).
We then use counterfactual variants to measure variant sensitivity (RVC) and counterfactual generalizability (CVA). Finally, we analyze how generalizability varies across reasoning types, diagnose failed updates, and illustrate these patterns with failure case studies. Table~\ref{tab:main-results} compiles our main results, as discussed below.

\vspace{-3pt}
\subsection{Reconstruction Controls}\label{sec:reconstruction-results}
\vspace{-3pt}

The group-average rows in Table~\ref{tab:main-results} show that RA is consistently close to or above OA across datasets, with the clearest increase on CharXiv.
This pattern likely reflects cleaner base reconstructions that increase resolution, reduce incidental clutter, or simplify graphics, while preserving task-relevant labels and relations. An example where the reconstruction is correctly answered despite the original failing is shown in
\Cref{fig:reconstruction-1} (see also \Cref{fig:reconstruction-2} of Appendix~\ref{app:reconstruction-example}). 
The two charts have low-level differences such as font choice, colors, and line shape that may contribute to VLM instability. 
Some original charts in the datasets have low resolution and visual artifacts because they were scraped from source documents, whereas our reconstructed charts are generated cleanly at high resolution from reverse-engineered code, which may benefit the VLM's ability to parse information.
We evaluate these possibilities directly. A human audit covers all 98 retained ChartMuseum charts and 142 chart-question families. All observed reconstruction artifacts are minor, and every family preserves the task-relevant evidence and remains answerable.
We also test the resolution hypothesis with five GPT-family models (\Cref{tab:resolution-ablation}). On average, reducing reconstruction resolution removes the full-resolution RA advantage over OA.

\begin{table}[t]
\centering
\footnotesize
\setlength{\tabcolsep}{3.2pt}
\resizebox{\columnwidth}{!}{\begin{tabular}{lrrrr}
\toprule
Model & OA & Full RA & RA (75\%) & RA (50\%) \\
\midrule
GPT-4.1 & 0.507 & 0.549 & 0.528 & 0.507 \\
GPT-4.1 mini & 0.528 & 0.528 & 0.486 & 0.521 \\
GPT-4o & 0.345 & 0.366 & 0.373 & 0.352 \\
GPT-5.4 & 0.704 & 0.718 & 0.683 & 0.662 \\
GPT-5.4 mini & 0.486 & 0.514 & 0.493 & 0.514 \\
\midrule
\textbf{Mean} & \textbf{0.514} & \textbf{0.535} & \textbf{0.513} & \textbf{0.511} \\
\bottomrule
\end{tabular}}
\caption{ChartMuseum reconstruction accuracy at full resolution and after resizing each image dimension to 75\% and 50\%.}
\label{tab:resolution-ablation}
\vspace{-8pt}
\end{table}

Finally, we recompute the counterfactual metrics using the reconstruction rather than the original chart as the control. The results remain close to the original-controlled results.
Together, these controls indicate that the generated charts preserve the benchmark tasks well enough for counterfactual comparison. The variant gap also remains under reconstruction-controlled evaluation. Full results are in Appendix~\ref{app:reconstruction-example}.

The per-dataset behavior diverges once counterfactual data changes are introduced.
ChartQA maintains high original and variant accuracy, suggesting that many conventional chart-question families transfer cleanly to the counterfactual setting. Every model shows a negative RVC, although the group-average declines are modest.
CharXiv remains difficult in absolute terms, yet its average VA is slightly higher than OA, yielding a small positive RVC that is not reliably different from zero.
However, this does not imply that originally solved CharXiv examples reliably generalize. Since VA pools families, gains on originally incorrect examples can offset losses on originally solved ones, which we discuss further in Section~\ref{subsec:cva_reveals}. 
ChartMuseum is different: RA stays close to OA, but VA drops, with statistically significant negative RVC for many models. This indicates VLMs can have trouble handling changed visual evidence distinct from reconstruction artifacts.
Open-source models follow the same trends as proprietary models, though they are lower in absolute performance across the harder benchmarks.

\vspace{-4pt}
\subsection{CVA Reveals Failed Generalization After Original-Chart Success}
\label{subsec:cva_reveals}
\vspace{-4pt}

\begin{figure}[t]
\centering
\includegraphics[width=\columnwidth, trim={7 8 8 7}, clip]{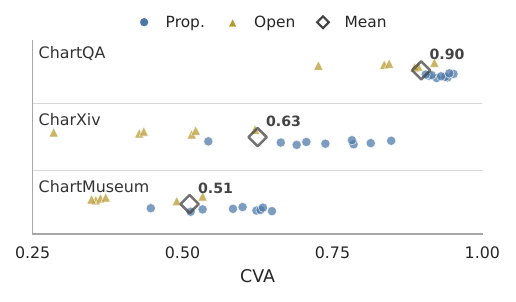}
\caption{Distribution of conditional variant accuracy (CVA) by dataset and model group. Points are individual models, colors and shapes distinguish proprietary and open-source models, and diamonds mark the mean over model CVA values within each dataset.}
\label{fig:generalization-after-success}
\vspace{-10pt}
\end{figure}

The central question we investigate is whether VLMs remain grounded on altered visual evidence after they have demonstrated competence on the original chart.
Aggregate VA and RVC are indicative, but do not directly address this.
Therefore, we use CVA (\Cref{sec:counterfactual-evaluation}) which restricts evaluation to families where the original chart was answered correctly, and then measures variant accuracy.
This conditioning isolates counterfactual generalizability on solvable questions.
Figure~\ref{fig:generalization-after-success} reports CVA by dataset and model group, while Appendix~\ref{sec:appendix-cva} reports the per-model values.

Proprietary models have higher CVA on average than open-source models, indicating better generalization.
Across datasets, ChartQA has high counterfactual generalizability, whereas CharXiv and ChartMuseum have substantially lower CVA.
CharXiv illustrates why the conditional view matters: its average RVC is not reliably negative, but CVA is substantially lower than 1.0.
ChartMuseum combines both signals, with a large negative RVC and even lower CVA.
Overall, VLMs are not consistently able to generalize to charts with altered visual evidence, even after solving the original.
We test robustness to inference stochasticity using the primary run and two additional independent runs. Of 321 ChartMuseum model--family failures across five GPT-family models, 216 (67.3\%) recur in both additional runs, allowing the failing variant to differ. VA remains below OA for every model in all three runs (Appendix~\ref{sec:appendix-repeated-inference}).
The following sections characterize where and how these failures arise.

\subsection{Generalizability Is Weakest for Visually Grounded Questions}\label{sec:reasoning-types}

\begin{figure}[t]
\centering
\includegraphics[width=\columnwidth, trim={5 10 12 10}, clip]{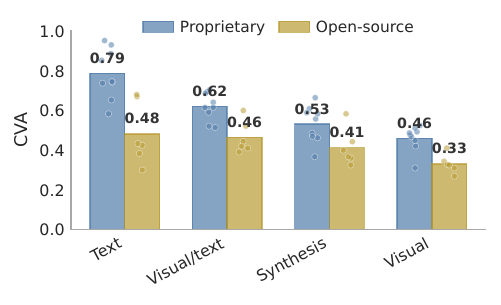}
\caption{ChartMuseum CVA by reasoning type and model group. Bars show mean CVA for each group, with individual model scores overlaid. Reasoning-type names follow the ChartMuseum terminology.}
\label{fig:chartmuseum-reasoning-types}
\vspace{-10pt}
\end{figure}

We next examine which chart-reasoning demands make counterfactual generalizability difficult.
ChartMuseum's reasoning-type annotations let us test whether generalizability varies across different kinds of chart reasoning.
Following ChartMuseum terminology~\cite{tang2026chartmuseumtestingvisualreasoning}, the categories distinguish Visual, Text, Visual/text, and Synthesis questions, depending on whether the answer is grounded in visual marks, explicit text on the chart, either source, or their combination.
Figure~\ref{fig:chartmuseum-reasoning-types} reports ChartMuseum CVA by category, and Appendix~\ref{sec:appendix-reasoning-types} gives the per-model breakdown.

The best generalization occurs for Text questions, followed by Visual/text questions, where explicit text can still support the answer. CVA is lower for Synthesis and Visual questions in both model groups, suggesting that counterfactual updating is particularly difficult when answers depend more directly on visual grounding.

\vspace{-2pt}
\subsection{Failed Generalization Manifests as Stale and Noisy Updates}
\label{sec:update-failure-analysis}
\vspace{-2pt}

When visual evidence changes after a successful solve, a VLM may answer the variant correctly (correct update, CU), repeat its original prediction incorrectly (stale prediction, SP), or change to another wrong answer (noisy update, NU). 
SP is consistent with failed re-grounding after the visual evidence changes and may reflect visual insensitivity or reliance on parametric knowledge of the question or chart from prior exposure during training.
NU can indicate an attempted update which is not grounded in visual evidence, or a reasoning failure.
Table~\ref{tab:update-failure-modes} displays the outcome rates for these three categories, aggregating over models (Appendix~\ref{sec:appendix-update-diagnostics} reports the per-model results).

\begin{table}[t]
\centering
\footnotesize
\setlength{\tabcolsep}{3pt}
\begin{tabular}{@{}lrrrrrr@{}}
\toprule
 & \multicolumn{3}{c}{Proprietary} & \multicolumn{3}{c}{Open-source} \\
\cmidrule(lr){2-4}\cmidrule(l){5-7}
Dataset & CU & NU & SP & CU & NU & SP \\
\midrule
ChartQA & \cellcolor{mygreen!18}0.929 & \cellcolor{mypurple!28}0.066 & \cellcolor{mypurple!65}\textcolor{white}{0.005} & \cellcolor{mygreen!18}0.851 & \cellcolor{mypurple!28}0.125 & \cellcolor{mypurple!65}\textcolor{white}{0.023} \\
CharXiv & \cellcolor{mygreen!18}0.726 & \cellcolor{mypurple!28}0.214 & \cellcolor{mypurple!65}\textcolor{white}{0.060} & \cellcolor{mygreen!18}0.473 & \cellcolor{mypurple!28}0.378 & \cellcolor{mypurple!65}\textcolor{white}{0.149} \\
ChartMuseum & \cellcolor{mygreen!18}0.579 & \cellcolor{mypurple!28}0.327 & \cellcolor{mypurple!65}\textcolor{white}{0.094} & \cellcolor{mygreen!18}0.410 & \cellcolor{mypurple!28}0.432 & \cellcolor{mypurple!65}\textcolor{white}{0.158} \\
\bottomrule
\end{tabular}
\caption{Update outcomes on counterfactual charts after success on the original, averaged within each model group. For each dataset and group, counterfactual variants are partitioned into correct updates (CU), noisy updates (NU), and stale predictions (SP).}
\label{tab:update-failure-modes}
\vspace{-10pt}
\end{table}

NU is the more common failure type, showing that VLMs often change their response without computing or grounding the updated answer correctly. SP is also substantial for the harder datasets CharXiv and ChartMuseum, especially for open-source models.
The prevalence of SP suggests that VLMs may sometimes rely on factors other than the presented evidence when answering questions, such as parametric knowledge of the question or chart.
To test whether SP cases reuse original-chart reasoning, we compare rationales from GPT-5.4, Claude Sonnet 4.6, and Gemini 2.5 Pro. For each model, we compare rationales from all SP cases with those from an equal-sized sample of NU cases. We measure lexical overlap using token-level ROUGE-L and use an independent LLM judge to assign a blinded 0--100 score indicating how strongly the variant rationale reuses chart-specific claims from the original rationale, following prior work on structured LLM-based evaluation~\cite{liu2023g}. Full sampling and judging details are provided in Appendix~\ref{sec:appendix-rationale-similarity-evaluation}; the complete similarity results are in Appendix~\ref{sec:appendix-rationale-similarity-results}.

As shown in \Cref{tab:rationale-similarity}, rationales from SP cases have consistently greater lexical overlap and substantially higher claim-reuse scores than those from NU cases across all three models; the semantic- and reasoning-similarity results in Appendix~\ref{sec:appendix-rationale-similarity-results} show the same pattern. The case studies in \Cref{sec:failure-case-studies} illustrate this quantitative pattern, which is consistent with some SP cases reusing original-chart information and with possible memorization. However, these behavioral and similarity analyses do not distinguish memorization from stochasticity, variant difficulty, visual misgrounding, or other reasoning errors.

\begin{figure*}[!t]
\centering
\includegraphics[width=\textwidth, trim={50 120 80 35}, clip]{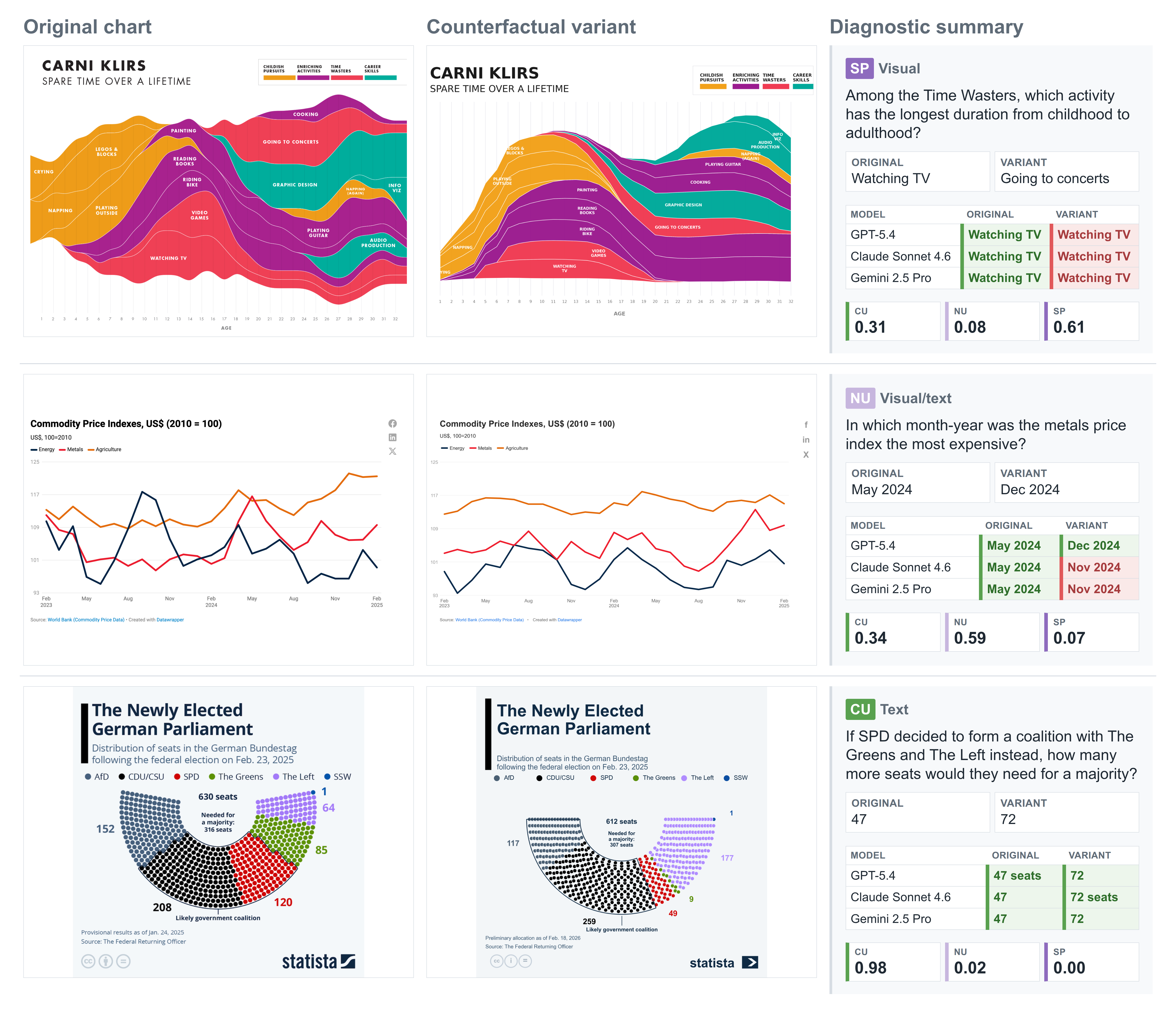}
\caption{ChartMuseum failure case studies illustrating update outcomes. Each row shows the original chart, one counterfactual variant, gold answers, and representative proprietary-model answers. Colors mark correct and incorrect predictions. Summary rates are family-level CU/NU/SP diagnostics over proprietary models and counterfactual variants. Additional generated rationales for the first two rows are analyzed in Appendix~\ref{sec:appendix-chartmuseum-case-studies}.}
\label{fig:failure-case-studies}
\vspace{-10pt}
\end{figure*}

\begin{table}[t]
\centering
\footnotesize
\setlength{\tabcolsep}{1.5pt}
\begin{tabular}{@{}l*{4}{>{\centering\arraybackslash}p{0.15\columnwidth}}@{}}
\toprule
& \multicolumn{2}{c}{ROUGE-L} & \multicolumn{2}{c}{Claim reuse} \\
\cmidrule(lr){2-3}\cmidrule(l){4-5}
Model & NU & SP & NU & SP \\
\midrule
GPT-5.4 & \cellcolor{mypurple!28}0.407 & \cellcolor{mypurple!65}\textcolor{white}{0.512} & \cellcolor{mypurple!28}59.2 & \cellcolor{mypurple!65}\textcolor{white}{89.1} \\
Claude Sonnet 4.6 & \cellcolor{mypurple!28}0.400 & \cellcolor{mypurple!65}\textcolor{white}{0.465} & \cellcolor{mypurple!28}57.8 & \cellcolor{mypurple!65}\textcolor{white}{86.8} \\
Gemini 2.5 Pro & \cellcolor{mypurple!28}0.389 & \cellcolor{mypurple!65}\textcolor{white}{0.454} & \cellcolor{mypurple!28}65.7 & \cellcolor{mypurple!65}\textcolor{white}{85.6} \\
\bottomrule
\end{tabular}
\caption{Rationale similarity on ChartMuseum after original-chart success. Mean token-level ROUGE-L and blinded claim-reuse scores (0--100) for all SP cases and equal-sized NU samples; higher values indicate greater similarity or reuse.}
\label{tab:rationale-similarity}
\vspace{-8pt}
\end{table}

\subsection{Failure Case Studies}
\label{sec:failure-case-studies}

Figure~\ref{fig:failure-case-studies} illustrates how models can produce a stale prediction, a noisy update, or a correct update after original-chart success when the visual evidence changes and the question remains fixed.
The case-study CU/NU/SP rates are computed over proprietary models only.
Motivated by the reasoning-type breakdown from \Cref{sec:reasoning-types}, we select a Visual case with a high SP rate, a Visual/text case with a high NU rate, and a Text case with a high CU rate for contrast.
Appendix~\ref{sec:appendix-charxiv-case-studies} extends the case studies to CharXiv.

\begin{figure}[t]
\centering
\begin{minipage}[t]{\columnwidth}
\textbf{Original Chart}
\begin{mytextbox}
\scriptsize\raggedright
The Time Wasters category is represented by the red/coral color in the chart.
Looking at the red-colored activities, I can identify: Video Games and Watching TV as the main Time Wasters (red color).
\textcolor{myblue}{Watching TV appears to start very early} (around age 1--2) \textcolor{myblue}{and continues throughout the entire lifetime shown (up to age 32) making it the longest duration Time Waster activity.}
Video Games appears to start around age 8--9 and continues into adulthood but seems to taper off.
\textcolor{myblue}{Watching TV spans from early childhood} (age 1--2) \textcolor{myblue}{all the way through adulthood} (age 32), \textcolor{myblue}{giving it the longest duration among the Time Wasters.}
\textbf{Answer:} Watching TV.
\end{mytextbox}

\vspace{2pt}
\textbf{Counterfactual Variant}
\begin{mytextbox}
\scriptsize\raggedright
Time Wasters are represented by the red/coral color in the chart.
Looking at the red-colored activities, I can identify:
1. \textcolor{red}{Watching TV -- appears to start around age 3--4 and continues through to age 32.}
2. Video Games -- appears from around age 8--9 and continues.
3. Going to Concerts -- appears from \textcolor{red}{around age 17--18} and continues.
\textcolor{red}{Watching TV (red) appears to have the longest duration}, \textcolor{red}{spanning from very early childhood} (around age 3--4) \textcolor{red}{all the way through to adulthood (age 32)}, \textcolor{red}{making it the longest-running Time Waster activity.}
Going to Concerts also appears in red and spans from about age 17 to 32.
\textcolor{red}{Watching TV clearly has the longest duration among Time Wasters}, \textcolor{red}{starting in early childhood and continuing through adulthood.}
\textbf{Answer:} Watching TV.
\end{mytextbox}
\end{minipage}
\caption{Claude Sonnet 4.6 responses for the first case study in \Cref{fig:failure-case-studies}. Highlighted spans show key reasoning points for the original (blue) and their incorrect reuse on the counterfactual variant (red). The counterfactual rationale better describes the original chart, a pattern consistent with original-chart information reuse and possible memorization.}
\label{fig:case1-claude-answer-comparison}
\vspace{-10pt}
\end{figure}

The first row shows stale predictions in a visual comparison after the correct answer changes: the displayed models keep their original predictions.
The answer requires comparing irregular band lengths along the age axis without explicit endpoint indicators, while the variant's similar overall shape makes the changed relations easy to miss. 
To further analyze this example, \Cref{fig:case1-claude-answer-comparison} shows Claude Sonnet 4.6's generated rationales for the original and variant.
This comparison illustrates the aggregate claim-reuse pattern in \Cref{tab:rationale-similarity}: the variant rationale is not grounded in the changed visual evidence and instead reuses key ``facts'' that better describe the original chart.
This pattern is consistent with possible memorization of the original chart or QA pair and, more generally, reliance on parametric knowledge rather than the displayed evidence.
More broadly, performance on published chart QA benchmarks may reflect memorization or other shortcuts alongside visual reasoning ability.
By generating new counterfactual chart-question families that preserve the task structure and are less likely to overlap with the evaluated models' training corpora, our method provides a more controlled measurement of visual reasoning under changed evidence.
Full analysis of the other models' generated rationales in this case study, and additional evidence consistent with possible memorization, are given in Appendix~\ref{sec:appendix-chartmuseum-case-studies}.

The second row shows a noisy update on an example that requires visual reasoning to locate the peak of a line, and textual reasoning to infer which month this peak occurs in.
In contrast to the original, the correct month label does not explicitly appear on the axis and needs to be inferred, making the counterfactual variant more challenging.
Only GPT-5.4 makes this update correctly. The other displayed VLMs select nearby text instead of inferring the missing month label.
This example demonstrates how counterfactual charts can introduce novel reasoning pathways not required in the original chart.

The third row displays clear party labels, distinguishable colors, and explicit seat annotations given in text. 
Here, models do not need to reason over graphical marks to reach a solution, but merely extract and associate text correctly before performing a simple calculation. 
All displayed models correctly update from 47 to 72.

Together, the cases contrast successful updating with two ways counterfactual generalization can fail: stale predictions and noisy updates.

%% file: sections/06_conclusion.tex
\section{Conclusion}

We introduced \textsc{Chartographer}, a framework for constructing counterfactual chart-question families that evaluate whether VLM chart QA remains grounded when chart data changes. 
Our pipeline first reverse engineers each chart into executable code through an iterative process with quality controls, updates the underlying data, and then re-renders variants that maintain the overall task.
Across ChartQA, CharXiv, and ChartMuseum, counterfactual variants expose failures that single-chart evaluation can hide.
Models may answer the original chart correctly yet fail the same task when the chart is re-rendered with altered data, with errors split between stale predictions and noisy updates.
These failures are especially visible when models must re-ground their reasoning on updated visual evidence, rather than rely on explicit textual cues.
Counterfactual evaluation therefore reveals generalizability gaps that single-chart performance can miss, exposes failures consistent with possible memorization or other reliance on parametric knowledge, and overall provides a new tool for diagnosing VLM chart reasoning.
Code, data-generation modules, and complete prompt templates are available at \url{https://github.com/compling-wat/Chartographer}.

%% file: sections/07_limitations.tex
\section*{Limitations}

Our study has three main limitations.
First, executable generation requires recoverable data, visual encodings, and question semantics. Although we retain 95.8\% of source charts, extremely dense plots and bespoke encodings remain outside the pipeline's coverage, so our conclusions may not extend to them.

Second, \textsc{Chartographer} isolates updating under changed answer-relevant data while preserving the chart schema and question task. It therefore does not measure robustness to changes in chart type, layout, visual encoding, or question formulation. Future work can extend the framework to these interventions.

Third, following ChartMuseum~\cite{tang2026chartmuseumtestingvisualreasoning}, we use an LLM judge rather than independent human validation to assess answer equivalence, which may introduce judge-specific errors or biases. Testing robustness to judge choice yields 98.05--98.21\% pairwise agreement and nearly identical aggregate metrics across three judges, though human validation remains desirable.

%% file: sections/A-appendix.tex
\section{Implementation Details}
\label{sec:appendix-implementation}
All prompts, judge instructions, and executable module templates described in this section are included in the released code.

\subsection{Source Data, Filtering, and Licenses}
\label{sec:appendix-source-data}

Table~\ref{tab:appendix-datasets} summarizes the source and filtered samples used in the experiments.
We exclude charts when chart-to-code reconstruction cannot support controlled variation, such as charts with distributions that cannot be clearly parameterized or extremely dense values that cannot be reliably recovered from the image.
We also exclude cases where the target relation depends on visual details that cannot be reconstructed reliably enough for controlled data edits.
The QA-pair count is also the number of accepted chart-question families before counterfactual expansion.

\begin{table}[h]
\centering
\footnotesize
\setlength{\tabcolsep}{4pt}
\resizebox{\columnwidth}{!}{\begin{tabular}{lrrrr}
\toprule
Dataset & \shortstack{Source\\charts} & \shortstack{Retained\\charts} & Retention & QA pairs \\
\midrule
ChartQA validation & 75 & 75 & 100.0\% & 150 \\
CharXiv validation & 150 & 148 & 98.7\% & 148 \\
ChartMuseum development & 110 & 98 & 89.1\% & 142 \\
\midrule
\textbf{Total} & \textbf{335} & \textbf{321} & \textbf{95.8\%} & \textbf{440} \\
\bottomrule
\end{tabular}}
\caption{Source coverage and filtered samples used for family construction. QA pairs correspond to accepted chart-question families before counterfactual expansion.}
\label{tab:appendix-datasets}
\end{table}

The two excluded CharXiv charts are highly dense plots whose underlying values cannot be recovered reliably from pixels. The twelve excluded ChartMuseum charts primarily use bespoke encodings that are difficult to reproduce with our plotting stack. Examples include Sankey-style energy-flow diagrams with many interconnected paths and radial layouts with custom angular geometry. The excluded cases reflect current executable-reconstruction limits, while the retained ChartMuseum set spans maps, timelines, scatter plots, and infographics.

We release the \ours\ generation pipeline under the Apache 2.0 license and our counterfactual chart-question families under the CC BY-SA 4.0 license where permitted by the source terms.
Derived examples remain subject to source licenses and provenance requirements: ChartQA is GPL-3.0,\footnote{\url{https://github.com/vis-nlp/ChartQA}} ChartMuseum is CC BY-SA 4.0 with chart copyrights retained by original sources,\footnote{\url{https://huggingface.co/datasets/lytang/ChartMuseum}} and CharXiv licenses original non-chart data under CC BY-SA 4.0 while retaining chart copyrights with the source papers.\footnote{\url{https://github.com/princeton-nlp/CharXiv}}
We preserve source attribution and release these families for use as an evaluation benchmark.

\subsection{Reconstruction}
\label{sec:appendix-reconstruction}

All model-assisted construction calls use GPT-5.2 (\texttt{gpt-5.2}) at temperature 0.
The reconstruction prompt requests semantic chart data and a Python \texttt{make\_figure(data, savepath=None)} function in JSON.
The data object stores entities, categories, labels, values, groups, and domain orderings, while code controls style and layout. Task-irrelevant visual elements may be abstracted when question-relevant evidence is preserved.

\subsection{Self-Improvement and Human Validation}
\label{sec:appendix-self-improvement-human-validation}

After reconstruction, automated diagnostics compare the source image and rendered reconstruction.
Cases with mismatches requiring correction enter self-improvement, which can proceed for up to five diagnostic and revision turns before the reconstruction is accepted, rejected, or sent for human review.
Four reviewers first assess a preliminary ChartMuseum sample independently to calibrate the criteria and examine agreement, then discuss disagreements to align the evaluation standards. The full collection is then divided among the reviewers, with one reviewer responsible for each assigned subset.
Because each item in the full review is labeled by one reviewer, we do not report full-dataset inter-annotator agreement.

Reviewers check layout, marks, scales, labels, legends, panel structure, styling, encoded values, and whether the target question remains answerable from the visible chart.
Human intervention is used for recurring problems that automated revision cannot resolve, such as ambiguous or overlapping labels, unreadable values, missing or inconsistent legends, incorrect panel or axis structure, mismatched mark types, or task-relevant relations that automated diagnostics cannot verify.
Charts are rejected when the reconstruction is too far from the source image, when the relevant visual relation cannot be recovered, or when the data schema cannot support controlled variation.
This review is task-focused: a reconstruction can be accepted if it preserves the visual evidence needed for the question, even when minor stylistic details differ from the source image. All reviewers are authors.
The systematic audit reported in \Cref{tab:appendix-reconstruction-audit} applies these criteria to all 98 retained ChartMuseum charts and their 142 chart-question families.

\subsection{Counterfactual Data Generation}
\label{sec:appendix-counterfactual-data-generation}

For each accepted chart, a generator prompt produces a deterministic Python \texttt{generate\_data(data\_template, seed)} module.
The generator must preserve the template schema, return valid same-schema data, and use seed-controlled regimes rather than only small noise.
For each accepted family, the base reconstruction keeps the original reconstructed data, while ten variants execute the family-specific generator with seeds 0--9.
The generator changes values, rankings, compositions, or trajectories enough to create meaningful counterfactual charts while preserving the original schema and chart semantics.
The generator does not assign fixed perturbation labels. One generated variant may alter several relations, and different seeds may produce different combinations.
Because these changes are non-mutually exclusive, we analyze outcomes by reasoning type and case study.
For charts requiring implicit context, assumption files record background knowledge, completeness constraints, and generation guidance.

\subsection{QA Generation}
\label{sec:appendix-qa-generation}

For each chart-question pair, the pipeline produces a Python \texttt{adapt\_question(data)} module.
The adapter normally returns the original question verbatim and makes only minimal validity edits when regenerated same-schema data would otherwise make the question unanswerable.
The \texttt{generate\_answer(data)} module computes the gold answer from the current data object rather than copied constants. It must use fields present in the schema, preserve the answer format implied by the question, and generalize to reordered elements and changed values.
Both modules are parsed, restricted to a single top-level function, and executed on the reconstructed data before use.

\subsection{Models}
\label{sec:appendix-models}

We use the same model set for all three datasets.
The proprietary group contains Claude Haiku 4.5 and Claude Sonnet 4.6, Gemini 2.5 Flash and Gemini 2.5 Pro, and GPT-4.1, GPT-4.1 mini, GPT-4o, GPT-5.4, and GPT-5.4 mini~\cite{anthropic2025haiku45,anthropic2026sonnet46,google2025gemini25models,openai2024gpt4o,openai2025gpt41,openai2026gpt54,openai2026gpt54mini}.
The open-source group contains Gemma 4 E4B IT, InternVL3-8B, LLaVA-OneVision-Qwen2-7B, Pixtral-12B, Qwen2.5-VL-7B-Instruct, and Qwen3-VL-8B-Instruct~\cite{google2026gemma4,zhu2025internvl3,li2024llavaonevision,agrawal2024pixtral12b,bai2025qwen3,bai2025qwen25vl}.
Proprietary models use hosted APIs; open-source models use local Hugging Face checkpoints with the same prediction and evaluation wrapper.
Local runs use a single NVIDIA H100 or L40S GPU with 128GB host memory, \texttt{bfloat16} inference, and a 1024-token generation limit by default.
The hosted Claude models use the same 1024-token output limit.
Hosted OpenAI and Gemini calls use provider defaults; local generation arguments other than precision and output length use checkpoint defaults.

\subsection{Prompting and Evaluation}
\label{sec:appendix-prompting-evaluation}

\paragraph{Model prompting.}
For every original chart, base reconstruction, and accepted counterfactual variant, the model receives only the chart image and the applicable question.
Gold answers, reconstructed data, generation seeds, executable code, and regime labels are never provided.
The fixed prompt is shown below. \texttt{[QUESTION]} is replaced by the applicable question.

\begin{mytextbox}
\texttt{Please answer the question using the chart image.\\[2pt]
Question: [QUESTION]\\[2pt]
Please first provide a brief rationale and then provide the answer. Use the following format:\\[2pt]
<rationale>\\
... your rationale here ...\\
</rationale>\\
<answer>\\
... your final answer (entity(s) or number) ...\\
</answer>}
\end{mytextbox}

\paragraph{Answer-equivalence judging.}
We score only the extracted final answer. If the tags are missing, we use the model's final response text.
For counterfactual variants, the question is the original question unless the adapter module makes a minimal validity edit.

The answer-equivalence judge receives the question, the target answer, and the extracted model answer, and returns a binary equivalence decision.
The judge ignores superficial formatting differences, capitalization differences, and unit formatting when they do not change the answer.
Numeric answers are judged by semantic equivalence rather than exact string identity when formatting differs without changing the value.
Exact calendar dates and explicitly labeled years must match the target answer, while nearby approximations are allowed for values read from an axis when they are visually indistinguishable at the chart resolution. Named entities, categorical labels, and ordinal choices must match the target answer.
When a question asks for multiple items, the prediction must contain the required set of items without adding contradictory answers.
The instantiated prompt has the following form: \texttt{Question: [QUESTION]}, \texttt{Answer 1: [TARGET]}, and \texttt{Answer 2: [PREDICTION]}, followed by an instruction to return exactly \texttt{Yes} or \texttt{No} without explanation.
We use GPT-4.1 mini through BespokeLabs Curator as the judge, run with temperature 0 and a 128-token output limit, and apply the same judging procedure to original charts, base reconstructions, and counterfactual variants.
The judge supplies only the binary correctness labels. CU, NU, and SP are subsequently derived deterministically from those labels and exact comparison of normalized extracted answers, as formalized in \Cref{sec:appendix-metric-definitions}.

\paragraph{Judge robustness.}
We evaluate robustness to judge choice by rescoring all 25,560 ChartMuseum decisions with Claude Haiku 4.5 and GPT-5.4 mini in addition to the primary GPT-4.1 mini judge.
Pairwise decision agreement is 98.05--98.21\%, and all three judges agree on 97.19\% of decisions.
As shown in \Cref{tab:appendix-judge-robustness}, the aggregate conclusions are insensitive to the judge choice.
All reported aggregate tables and diagnostics are computed from the final prediction and evaluation files using this primary judging configuration.

\begin{table}[t]
\centering
\footnotesize
\setlength{\tabcolsep}{3.5pt}
\resizebox{\columnwidth}{!}{\begin{tabular}{lrrrrr}
\toprule
Judge & OA & RA & VA & RVC & CVA \\
\midrule
Claude Haiku 4.5 & 0.455 & 0.448 & 0.360 & $-19.43\%$ & 0.513 \\
GPT-4.1 mini & 0.450 & 0.449 & 0.362 & $-18.18\%$ & 0.512 \\
GPT-5.4 mini & 0.458 & 0.450 & 0.362 & $-19.52\%$ & 0.511 \\
\bottomrule
\end{tabular}}
\caption{ChartMuseum metrics averaged over the 15 evaluated models after rescoring all predictions with three answer-equivalence judges.}
\label{tab:appendix-judge-robustness}
\end{table}

\paragraph{Rationale-similarity evaluation.}
\label{sec:appendix-rationale-similarity-evaluation}
For GPT-5.4, Claude Sonnet 4.6, and Gemini 2.5 Pro, we include every stale prediction and an equal-sized random sample of noisy updates, giving 91, 65, and 96 examples per outcome, respectively. Sampling uses seed 42 for GPT-5.4 and Claude Sonnet 4.6 and seed 43 for Gemini 2.5 Pro, yielding 504 model--variant comparisons from 101 ChartMuseum families. Token-level ROUGE-L measures lexical overlap between the original and variant rationales. GPT-4.1 mini receives the question and both rationales, but not the answers, correctness judgments, or behavioral outcome labels. It assigns three scores from 0 to 100: semantic similarity measures whether the rationales express the same propositions and conclusion, reasoning-strategy similarity measures whether they follow the same chart-reading or inference procedure, and chart-specific claim reuse measures whether factual claims from the original rationale reappear in the variant rationale. The judge ignores generic overlap induced by the shared question or standard chart-reading language, is independent of the three response models, and runs at temperature 0.

\paragraph{Resolution ablation.}
For the resolution ablation, each ChartMuseum reconstruction is converted to RGB, resized to 75\% and 50\% of its original width and height using bicubic interpolation, and saved as a PNG. Models otherwise use the same questions, prompts, and evaluation procedure as in the full-resolution condition.

\subsection{Metric Definitions}
\label{sec:appendix-metric-definitions}

For dataset $d$, let $F_d$ index its accepted source QA pairs. After counterfactual expansion, each $i \in F_d$ defines one chart-question family.
Let $V_i$ denote the seed set used to generate variants for index $i$. In our experiments, $V_i=\{0,\ldots,9\}$ for all $i$.
For model $m$ and source QA pair $i$, let $y_i^o$, $y_i^r$, and $y_{ij}^v$ be the gold answers for the original chart, base reconstruction, and variant seed $j$.
Let $\hat{y}_{m,i}^o$, $\hat{y}_{m,i}^r$, and $\hat{y}_{m,ij}^v$ be the corresponding model predictions.
Here, $\equiv$ denotes judged answer equivalence.

Original accuracy (OA), reconstruction accuracy (RA), and variant accuracy (VA) are
\[
\mathrm{OA}_{m,d}
= \frac{1}{|F_d|}
\sum_{i \in F_d} \mathbf{1}[\hat{y}_{m,i}^o \equiv y_i^o],
\]
\[
\mathrm{RA}_{m,d}
= \frac{1}{|F_d|}
\sum_{i \in F_d} \mathbf{1}[\hat{y}_{m,i}^r \equiv y_i^r],
\]
\[
\mathrm{VA}_{m,d}
=
\frac{
\sum_{i \in F_d}\sum_{j \in V_i}
\mathbf{1}[\hat{y}_{m,ij}^v \equiv y_{ij}^v]
}{
\sum_{i\in F_d}|V_i|
}.
\]
Relative variant change (RVC) is
\[
\mathrm{RVC}_{m,d} = 100\cdot\frac{\mathrm{VA}_{m,d}-\mathrm{OA}_{m,d}}{\mathrm{OA}_{m,d}}.
\]
Conditional variant accuracy (CVA) restricts VA to families where model $m$ answered the original chart correctly.
Let
\[
S_{m,d}=\{i \in F_d: \hat{y}_{m,i}^o \equiv y_i^o\}.
\]
Then
\[
\mathrm{CVA}_{m,d}
=
\frac{
\sum_{i \in S_{m,d}}\sum_{j \in V_i}
\mathbf{1}[\hat{y}_{m,ij}^v \equiv y_{ij}^v]
}{
\sum_{i\in S_{m,d}}|V_i|
}.
\]
This conditioning separates failure to solve the original task from failure to generalize after original-chart success.

Aggregate values reported in the main text and figures are computed from these per-model quantities.
For a model group $G$ such as proprietary or open-source models, we report the arithmetic mean over models:
\[
\mathrm{Metric}_{G,d}
=\frac{1}{|G|}\sum_{m\in G}\mathrm{Metric}_{m,d}.
\]
Dataset-level aggregate values use the same unweighted mean over all evaluated models.
For example, the aggregate RA/OA pairs and dataset-level CVA values in Section~\ref{sec:results} average the corresponding per-model metrics, rather than pooling all model predictions into a single denominator.

Update diagnostics use the same conditioning on original-chart success as CVA and include all generated seed variants, because every seed changes the visual evidence even when the answer remains the same.
Correctness against gold answers uses the judged answer-equivalence labels.
Whether a model changes its prediction is determined by exact match after final-answer extraction and normalization. Let $n(\cdot)$ denote this normalized answer string.
The restriction is applied separately for each model:
\[
D_{m,d}=\{(i,j): i\in F_d,\; j\in V_i,\; \hat{y}_{m,i}^o \equiv y_i^o\}.
\]
Because the original-success condition depends on the model prediction, $D_{m,d}$ can differ across models even within the same dataset.
Thus, a variant may contribute to the update diagnostics for one model but not for another.
With ten seed variants per expanded family, $|D_{m,d}|=10|S_{m,d}|$.
Within each model-specific set $D_{m,d}$, the update rates are
\[
\mathrm{CU}_{m,d}
=
\frac{
\sum_{(i,j)\in D_{m,d}}
\mathbf{1}[\hat{y}_{m,ij}^v \equiv y_{ij}^v]
}{
|D_{m,d}|
},
\]
\[
\mathrm{SP}_{m,d}
=
\frac{
\sum_{(i,j)\in D_{m,d}}
\mathbf{1}\!\left[
\begin{array}{@{}c@{}}
n(\hat{y}_{m,ij}^v) = n(\hat{y}_{m,i}^o) \\
\land\; \hat{y}_{m,ij}^v \not\equiv y_{ij}^v
\end{array}
\right]
}{
|D_{m,d}|
},
\]
\[
\mathrm{NU}_{m,d}
=
\frac{
\sum_{(i,j)\in D_{m,d}}
\mathbf{1}\!\left[
\begin{array}{@{}c@{}}
n(\hat{y}_{m,ij}^v) \ne n(\hat{y}_{m,i}^o) \\
\land\; \hat{y}_{m,ij}^v \not\equiv y_{ij}^v
\end{array}
\right]
}{
|D_{m,d}|
}.
\]
CU is the fraction of variants where the model predicts the variant answer correctly.
SP is the fraction where the model repeats its original prediction and is wrong on the variant.
NU is the fraction where the model changes away from its original prediction but is still wrong on the variant.
They form a partition of all seed variants after model-specific original-chart success and sum to one for each model--dataset aggregate.
These categories describe observed prediction behavior rather than the causal mechanism behind the error.
When update diagnostics are summarized by model group or dataset in the main text, we average the per-model CU, SP, and NU rates over the relevant model set.
This keeps each model equally weighted even though the model-specific diagnostic set $D_{m,d}$ can have different size.

\section{Additional Results}
\label{sec:appendix-additional-results}

\begin{figure*}[t]
    \centering
    \begin{minipage}[t]{0.48\textwidth}
        \centering
        \includegraphics[width=\textwidth]{}
        \begin{flushleft}
            \small
            \textbf{Question:} At which tie flow value (MW) is the difference between \( \dot{W}_{gr} \) and \( \dot{W}_{fr} \) the smallest?\\ \textbf{Gold:} 200 MW\\
            \textbf{GPT-5.4 mini:} The difference between \( \dot{W}_{gr} \) and \( \dot{W}_{fr} \) is the vertical gap between the gray solid curve and the black dashed curve. \textcolor{red}{Scanning across the tie flow axis, that gap is smallest at the far right end of the plot, where the two curves nearly meet.} \textbf{Answer: About 440 MW}
        \end{flushleft}
    \end{minipage}
    \hfill
    \vrule{} 
    \hfill
    \begin{minipage}[t]{0.48\textwidth}
        \centering
        \includegraphics[width=1\textwidth]{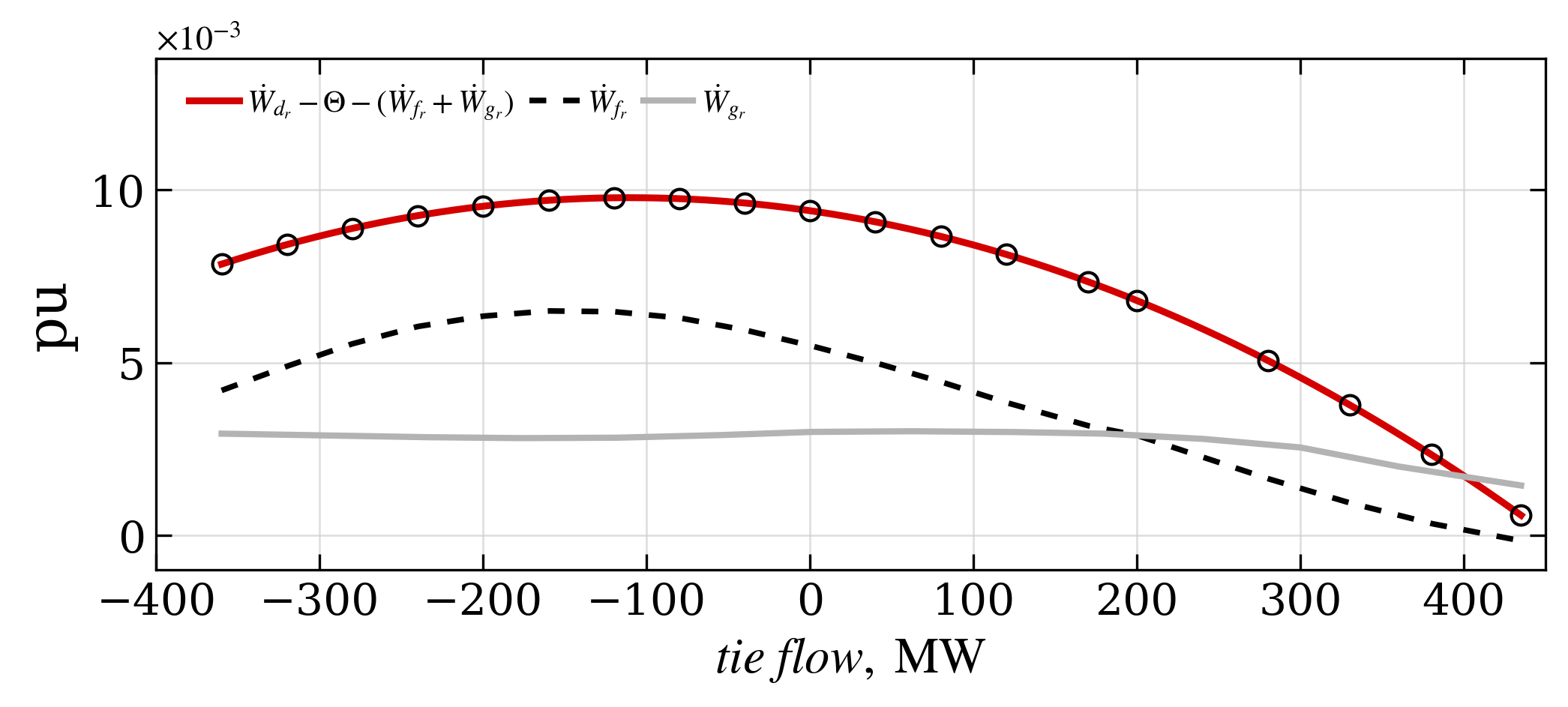}
        \begin{flushleft}
            \small
            \textbf{Question:} At which tie flow value (MW) is the difference between \( \dot{W}_{gr} \) and \( \dot{W}_{fr} \) the smallest?\\ \textbf{Gold:} 200 MW\\
            \textbf{GPT-5.4 mini:} The difference between \( \dot{W}_{gr} \) and \( \dot{W}_{fr} \) is smallest where the two curves are closest on the chart. Visually, they come closest around a tie flow of about 200 MW. \textbf{Answer: 200 MW}
        \end{flushleft}
    \end{minipage}
    
    \vspace{0.75em}
    
    \begin{minipage}[t]{0.48\textwidth}
        \centering
        \includegraphics[width=\textwidth]{}
        \begin{flushleft}
            \small
            \textbf{Question:} In which activity does the B-ResNet-TTT model outperform the J-ResNet model by the largest margin?\\ \textbf{Gold:} front pocket\\
            \textbf{GPT-5.4 mini:} To find where B-ResNet-TTT outperforms J-ResNet by the largest margin, compare the green bar against the orange bar for each activity. \textcolor{red}{The biggest gap where the green bar is lower than the orange bar appears in both ``calling'' and ``back pocket.''} \textbf{Answer: calling and back pocket}
        \end{flushleft}
    \end{minipage}
    \hfill
    \vrule{} 
    \hfill
    \begin{minipage}[t]{0.48\textwidth}
        \centering
        \includegraphics[width=1\textwidth]{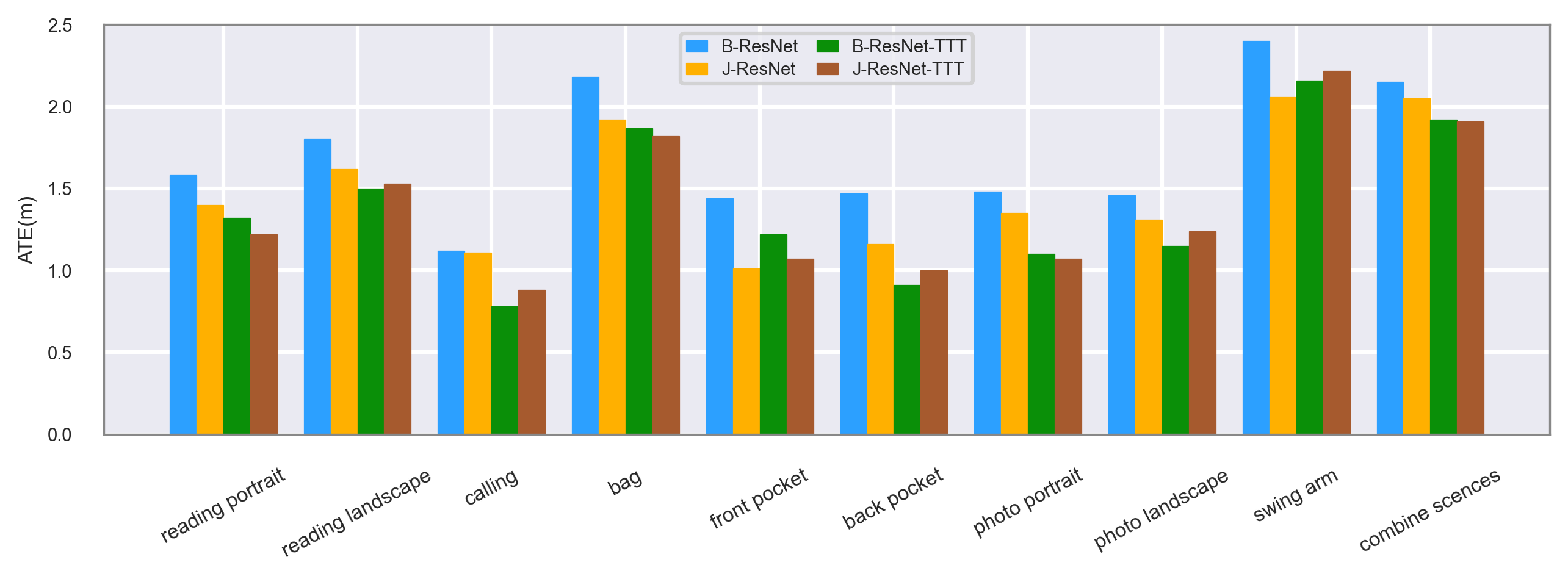}
        \begin{flushleft}
            \small
            \textbf{Question:} In which activity does the B-ResNet-TTT model outperform the J-ResNet model by the largest margin?\\ \textbf{Gold:} front pocket\\
            \textbf{GPT-5.4 mini:} Compare the B-ResNet-TTT bar to the J-ResNet bar for each activity and find where the gap is largest. The largest positive gap appears in front pocket. \textbf{Answer: front pocket}
        \end{flushleft}
    \end{minipage}
    \caption{Additional examples where GPT-5.4 mini gives a correct answer on the reconstruction (right), but not the original (left). The reconstruction has fewer artifacts and higher resolution because it was generated cleanly, not scraped from public sources. Flawed observations are shown in \textcolor{red}{red text}.}\label{fig:reconstruction-2}
\end{figure*}

\subsection{Reconstruction Accuracy Analysis}\label{app:reconstruction-example}

\paragraph{Qualitative examples.}
Our results on reconstructed charts from \Cref{sec:reconstruction-results} showed that VLMs had nearly equal, and sometimes superior, performance on tasks using the reconstructed charts compared to the originals. \Cref{fig:reconstruction-1} and \Cref{fig:reconstruction-2} show examples of this behavior where GPT-5.4 mini correctly answers an identical question on the reconstruction, but not the original. To a human, the two charts appear nearly identical in content, with only minor stylistic differences. However, to a VLM these changes may matter. Low-level differences such as font choice, colors, and line thickness may contribute to VLM instability, but we also note there can be a more significant difference in the resolution of original and reconstructed charts. Original charts from the ChartQA, CharXiv, and ChartMuseum datasets can be low-resolution because they were scraped from a variety of sources. In contrast, our reconstructed charts are generated cleanly at high resolution from reverse-engineered code, which may benefit a VLM's ability to parse information.

\paragraph{Human reconstruction audit.}
We audit reconstruction quality for all 98 retained ChartMuseum charts and all 142 associated chart-question families. Table~\ref{tab:appendix-reconstruction-audit} summarizes this finding.
Every chart has minor stylistic differences, but none has a major artifact. All 142 families preserve the task-relevant evidence and remain answerable. Of these families, 138 are judged similarly difficult after reconstruction, with two easier and two harder. The audit finds no systematic shift in task difficulty after reconstruction.

\begin{table}[t]
\centering
\footnotesize
\setlength{\tabcolsep}{4pt}
\begin{tabular}{llrr}
\toprule
Assessment & Category & Count & Percentage \\
\midrule
Artifact severity & Minor & 98 & 100.0\% \\
 & Major & 0 & 0.0\% \\
Evidence preserved & Yes & 142 & 100.0\% \\
Relative difficulty & Easier & 2 & 1.4\% \\
 & Similar & 138 & 97.2\% \\
 & Harder & 2 & 1.4\% \\
Answerable & Yes & 142 & 100.0\% \\
\bottomrule
\end{tabular}
\caption{Human audit of retained ChartMuseum reconstructions. Artifact severity is assessed per chart; all other criteria are assessed per QA family.}
\label{tab:appendix-reconstruction-audit}
\end{table}

\paragraph{Reconstruction-controlled evaluation.}
We also condition on reconstruction rather than original-chart success. CVA-R restricts variant accuracy to families whose reconstruction is answered correctly, while RVC-R uses reconstruction accuracy as its baseline and is computed as $100\cdot(\mathrm{VA}-\mathrm{RA})/\mathrm{RA}$. \Cref{tab:reconstruction-controlled} reports arithmetic means of the per-model metrics over the 15 models. The gap remains under this control, and the results are close to the original-controlled results.

\begin{table}[t]
\centering
\footnotesize
\setlength{\tabcolsep}{2pt}
\begin{tabular}{@{}lrrrrr@{}}
\toprule
Dataset & \shortstack{OA--RA\\agree.} & CVA-O & CVA-R & RVC-O & RVC-R \\
\midrule
ChartQA & 0.917 & 0.898 & 0.902 & $-4.3\%$ & $-4.5\%$ \\
CharXiv & 0.783 & 0.625 & 0.637 & $+1.1\%$ & $-6.1\%$ \\
ChartMuseum & 0.799 & 0.512 & 0.528 & $-18.2\%$ & $-17.1\%$ \\
\bottomrule
\end{tabular}
\caption{Metrics using the original and reconstruction controls, averaged over models. CVA-O is the primary CVA reported in the main paper. CVA-R conditions on reconstruction success.}
\label{tab:reconstruction-controlled}
\end{table}

The OA--RA agreement values in \Cref{tab:reconstruction-controlled} correspond to the pooled correctness counts in \Cref{tab:appendix-original-reconstruction-contingency}. O and R denote correctness on the original and reconstruction, respectively.

\begin{table}[t]
\centering
\footnotesize
\setlength{\tabcolsep}{4pt}
\begin{tabular}{lrrrr}
\toprule
Dataset & O$-$R$-$ & O$-$R$+$ & O$+$R$-$ & O$+$R$+$ \\
\midrule
ChartQA & 170 & 95 & 91 & 1,894 \\
CharXiv & 829 & 289 & 192 & 910 \\
ChartMuseum & 958 & 213 & 215 & 744 \\
\bottomrule
\end{tabular}
\caption{Original--reconstruction correctness counts pooled over the 15 evaluated models.}
\label{tab:appendix-original-reconstruction-contingency}
\end{table}

\subsection{Conditional Variant Accuracy}
\label{sec:appendix-cva}

Table~\ref{tab:appendix-cva-by-model} reports per-model CVA for each dataset.
It complements Table~\ref{tab:main-results} without repeating the OA, RA, VA, and RVC columns already reported in the main text.
These values provide the model-level spread shown in Figure~\ref{fig:generalization-after-success}.

\begin{table}[!t]
\centering
\footnotesize
\setlength{\tabcolsep}{4pt}
\resizebox{\columnwidth}{!}{\begin{tabular}{@{}lrrr@{}}
\toprule
Model & ChartQA & CharXiv & ChartMuseum \\
\midrule
\multicolumn{4}{l}{\textit{Proprietary models}} \\
Claude Haiku 4.5 & 0.924 & 0.686 & 0.513 \\
Claude Sonnet 4.6 & 0.942 & 0.791 & 0.649 \\
Gemini 2.5 Flash & 0.938 & 0.732 & 0.623 \\
Gemini 2.5 Pro & 0.931 & 0.801 & 0.630 \\
GPT-4.1 & 0.911 & 0.646 & 0.533 \\
GPT-4.1 mini & 0.915 & 0.698 & 0.584 \\
GPT-4o & 0.905 & 0.544 & 0.447 \\
GPT-5.4 & 0.952 & 0.848 & 0.634 \\
GPT-5.4 mini & 0.945 & 0.788 & 0.600 \\
\textbf{Proprietary average} & \textbf{0.929} & \textbf{0.726} & \textbf{0.579} \\
\midrule
\multicolumn{4}{l}{\textit{Open-source models}} \\
Gemma 4 E4B IT & 0.887 & 0.548 & 0.490 \\
InternVL3-8B & 0.893 & 0.428 & 0.355 \\
LLaVA-OneVision-Qwen2-7B & 0.726 & 0.276 & 0.348 \\
Pixtral-12B & 0.836 & 0.436 & 0.363 \\
Qwen2.5-VL-7B-Instruct & 0.844 & 0.508 & 0.372 \\
Qwen3-VL-8B-Instruct & 0.920 & 0.641 & 0.533 \\
\textbf{Open-source average} & \textbf{0.851} & \textbf{0.473} & \textbf{0.410} \\
\bottomrule
\end{tabular}
}
\caption{Conditional variant accuracy (CVA) by dataset and model. CVA is variant accuracy restricted to families where the model answered the original chart correctly.}
\label{tab:appendix-cva-by-model}
\end{table}

\subsection{Repeated-Inference Robustness}
\label{sec:appendix-repeated-inference}

We run each original and all ten variants for every ChartMuseum family with five GPT-family models, collecting two additional independent responses per input beyond the primary evaluation. We consider a failure from the primary evaluation reproduced when the original is correct in all three runs and every run contains at least one wrong variant. The failing seed may differ across runs. \Cref{tab:appendix-repeated-inference} shows that a majority of the initially observed failures are reproduced for every model. In total, 216 of 321 model--family cases (67.3\%) are reproduced. Individual failure membership is therefore stochastic, but the aggregate result is stable: VA remains below OA for every model in all three runs.

\begin{table}[t]
\centering
\footnotesize
\setlength{\tabcolsep}{3.5pt}
\begin{tabular}{lrrr}
\toprule
Model & Initial failures & Reproduced & Rate \\
\midrule
GPT-4.1 & 65 & 47 & 72.3\% \\
GPT-4.1 mini & 67 & 42 & 62.7\% \\
GPT-4o & 47 & 25 & 53.2\% \\
GPT-5.4 & 81 & 62 & 76.5\% \\
GPT-5.4 mini & 61 & 40 & 65.6\% \\
\midrule
\textbf{Total} & \textbf{321} & \textbf{216} & \textbf{67.3\%} \\
\bottomrule
\end{tabular}
\caption{Reproduction of ChartMuseum original-correct/variant-wrong model--family cases across three independent inference runs.}
\label{tab:appendix-repeated-inference}
\end{table}

\Cref{tab:appendix-repeated-metrics} compares the primary evaluation with the mean and standard deviation across all three runs: the primary run and two additional independent runs. The aggregate OA, VA, and CVA values remain close to the primary results.

\begin{table*}[t]
\centering
\footnotesize
\setlength{\tabcolsep}{3.5pt}
\begin{tabular}{lrrrrrr}
\toprule
& \multicolumn{2}{c}{OA} & \multicolumn{2}{c}{VA} & \multicolumn{2}{c}{CVA} \\
\cmidrule(lr){2-3}\cmidrule(lr){4-5}\cmidrule(l){6-7}
Model & Initial & 3-run & Initial & 3-run & Initial & 3-run \\
\midrule
GPT-4.1 & 0.507 & $0.488 \pm 0.027$ & 0.392 & $0.409 \pm 0.015$ & 0.533 & $0.552 \pm 0.028$ \\
GPT-4.1 mini & 0.528 & $0.552 \pm 0.020$ & 0.446 & $0.445 \pm 0.008$ & 0.584 & $0.564 \pm 0.020$ \\
GPT-4o & 0.345 & $0.364 \pm 0.033$ & 0.287 & $0.307 \pm 0.019$ & 0.447 & $0.463 \pm 0.020$ \\
GPT-5.4 & 0.704 & $0.690 \pm 0.014$ & 0.554 & $0.558 \pm 0.006$ & 0.634 & $0.667 \pm 0.029$ \\
GPT-5.4 mini & 0.486 & $0.505 \pm 0.016$ & 0.425 & $0.423 \pm 0.013$ & 0.600 & $0.599 \pm 0.019$ \\
\bottomrule
\end{tabular}
\caption{Primary and repeated-inference ChartMuseum metrics. Three-run columns report mean $\pm$ standard deviation over the primary run and two additional independent runs.}
\label{tab:appendix-repeated-metrics}
\end{table*}

\subsection{Update Diagnostics}
\label{sec:appendix-update-diagnostics}

Table~\ref{tab:appendix-update-by-model} reports per-model update outcomes on seed variants after original-chart success.
Each row is computed with that model's own diagnostic set $D_{m,d}$ for each dataset. The table should therefore be read as model-specific conditional rates rather than rates over a single shared pool of variants.
The table expands the group-level CU, NU, and SP patterns summarized in \Cref{tab:update-failure-modes}.

\begin{table*}[!t]
\centering
\footnotesize
\setlength{\tabcolsep}{4pt}
\resizebox{0.9\textwidth}{!}{\begin{tabular}{lrrrrrrrrr}
\toprule
 & \multicolumn{3}{c}{ChartQA} & \multicolumn{3}{c}{CharXiv} & \multicolumn{3}{c}{ChartMuseum} \\
\cmidrule(lr){2-4}\cmidrule(lr){5-7}\cmidrule(lr){8-10}
Model & CU & NU & SP & CU & NU & SP & CU & NU & SP \\
\midrule
\multicolumn{10}{l}{\textit{Proprietary models}} \\
Claude Haiku 4.5 & 0.924 & 0.074 & 0.003 & 0.686 & 0.266 & 0.048 & 0.513 & 0.413 & 0.073 \\
Claude Sonnet 4.6 & 0.942 & 0.054 & 0.004 & 0.791 & 0.168 & 0.041 & 0.649 & 0.285 & 0.066 \\
Gemini 2.5 Flash & 0.938 & 0.059 & 0.004 & 0.732 & 0.220 & 0.048 & 0.623 & 0.279 & 0.098 \\
Gemini 2.5 Pro & 0.931 & 0.066 & 0.003 & 0.801 & 0.154 & 0.045 & 0.630 & 0.278 & 0.092 \\
GPT-4.1 & 0.911 & 0.080 & 0.009 & 0.646 & 0.281 & 0.073 & 0.533 & 0.339 & 0.128 \\
GPT-4.1 mini & 0.915 & 0.078 & 0.007 & 0.698 & 0.215 & 0.087 & 0.584 & 0.355 & 0.061 \\
GPT-4o & 0.905 & 0.088 & 0.007 & 0.544 & 0.359 & 0.097 & 0.447 & 0.418 & 0.135 \\
GPT-5.4 & 0.952 & 0.045 & 0.004 & 0.848 & 0.121 & 0.031 & 0.634 & 0.275 & 0.091 \\
GPT-5.4 mini & 0.945 & 0.048 & 0.008 & 0.788 & 0.144 & 0.068 & 0.600 & 0.300 & 0.100 \\
\textbf{Proprietary average} & \textbf{0.929} & \textbf{0.066} & \textbf{0.005} & \textbf{0.726} & \textbf{0.214} & \textbf{0.060} & \textbf{0.579} & \textbf{0.327} & \textbf{0.094} \\
\midrule
\multicolumn{10}{l}{\textit{Open-source models}} \\
Gemma 4 E4B IT & 0.887 & 0.103 & 0.010 & 0.548 & 0.352 & 0.100 & 0.490 & 0.393 & 0.117 \\
InternVL3-8B & 0.893 & 0.067 & 0.040 & 0.428 & 0.341 & 0.230 & 0.355 & 0.463 & 0.182 \\
LLaVA-OneVision-Qwen2-7B & 0.726 & 0.235 & 0.039 & 0.276 & 0.535 & 0.189 & 0.348 & 0.381 & 0.270 \\
Pixtral-12B & 0.836 & 0.147 & 0.017 & 0.436 & 0.416 & 0.147 & 0.363 & 0.509 & 0.129 \\
Qwen2.5-VL-7B-Instruct & 0.844 & 0.129 & 0.026 & 0.508 & 0.378 & 0.114 & 0.372 & 0.474 & 0.154 \\
Qwen3-VL-8B-Instruct & 0.920 & 0.071 & 0.009 & 0.641 & 0.246 & 0.113 & 0.533 & 0.372 & 0.094 \\
\textbf{Open-source average} & \textbf{0.851} & \textbf{0.125} & \textbf{0.023} & \textbf{0.473} & \textbf{0.378} & \textbf{0.149} & \textbf{0.410} & \textbf{0.432} & \textbf{0.158} \\
\bottomrule
\end{tabular}
}
\caption{Update outcome rates by dataset and model after original-chart success. Seed variants are decomposed into correct update (CU), noisy update (NU), and stale prediction (SP).}
\label{tab:appendix-update-by-model}
\end{table*}

\paragraph{Rationale-similarity results.}
\label{sec:appendix-rationale-similarity-results}
Table~\ref{tab:appendix-rationale-similarity} reports the full rationale-similarity results using the evaluation procedure in Appendix~\ref{sec:appendix-rationale-similarity-evaluation}. Across all three models, SP rationales have higher ROUGE-L, semantic similarity, reasoning similarity, and chart-specific claim reuse than the equal-sized NU samples. This pattern is consistent with some SP cases reusing original-chart information, although it does not by itself distinguish memorization from other causes of failed re-grounding.

\begin{table}[t]
\centering
\footnotesize
\setlength{\tabcolsep}{3pt}
\begin{tabular}{@{}lrrrrr@{}}
\toprule
Outcome & $N$ & R-L & Sem. & Reas. & Claims \\
\midrule
\multicolumn{6}{@{}l}{\textbf{GPT-5.4}} \\
\quad SP & 91 & 0.512 & 89.1 & 93.1 & 89.1 \\
\quad NU & 91 & 0.407 & 57.8 & 80.5 & 59.2 \\
\addlinespace[2pt]
\multicolumn{6}{@{}l}{\textbf{Claude Sonnet 4.6}} \\
\quad SP & 65 & 0.465 & 84.9 & 88.8 & 86.8 \\
\quad NU & 65 & 0.400 & 57.7 & 72.2 & 57.8 \\
\addlinespace[2pt]
\multicolumn{6}{@{}l}{\textbf{Gemini 2.5 Pro}} \\
\quad SP & 96 & 0.454 & 86.8 & 90.2 & 85.6 \\
\quad NU & 96 & 0.389 & 62.9 & 80.2 & 65.7 \\
\bottomrule
\end{tabular}
\caption{Lexical and blinded rationale similarity between original and variant responses. R-L denotes ROUGE-L; Sem., Reas., and Claims denote semantic similarity, reasoning similarity, and chart-specific claim reuse (0--100).}
\label{tab:appendix-rationale-similarity}
\end{table}

\subsection{ChartMuseum Reasoning-Type Analysis}
\label{sec:appendix-reasoning-types}

\begin{table}[!t]
\centering
\footnotesize
\setlength{\tabcolsep}{4pt}
\resizebox{\columnwidth}{!}{\begin{tabular}{@{}lrrrr@{}}
\toprule
Model & Text & Visual/text & Synthesis & Visual \\
\midrule
\multicolumn{5}{l}{\textit{Proprietary models}} \\
Claude Haiku 4.5 & 0.653 & 0.514 & 0.557 & 0.421 \\
Claude Sonnet 4.6 & 0.930 & 0.695 & 0.486 & 0.524 \\
Gemini 2.5 Flash & 0.886 & 0.641 & 0.608 & 0.475 \\
Gemini 2.5 Pro & 0.952 & 0.615 & 0.664 & 0.468 \\
GPT-4.1 & 0.738 & 0.521 & 0.463 & 0.448 \\
GPT-4.1 mini & 0.742 & 0.616 & 0.470 & 0.493 \\
GPT-4o & 0.583 & 0.591 & 0.367 & 0.310 \\
GPT-5.4 & 0.852 & 0.705 & 0.583 & 0.511 \\
GPT-5.4 mini & 0.747 & 0.687 & 0.589 & 0.487 \\
\textbf{Proprietary average} & \textbf{0.787} & \textbf{0.620} & \textbf{0.532} & \textbf{0.460} \\
\midrule
\multicolumn{5}{l}{\textit{Open-source models}} \\
Gemma 4 E4B IT & 0.670 & 0.600 & 0.443 & 0.344 \\
InternVL3-8B & 0.425 & 0.444 & 0.360 & 0.269 \\
LLaVA-OneVision-Qwen2-7B & 0.300 & 0.391 & 0.367 & 0.310 \\
Pixtral-12B & 0.383 & 0.420 & 0.400 & 0.328 \\
Qwen2.5-VL-7B-Instruct & 0.433 & 0.410 & 0.325 & 0.325 \\
Qwen3-VL-8B-Instruct & 0.680 & 0.521 & 0.583 & 0.411 \\
\textbf{Open-source average} & \textbf{0.482} & \textbf{0.464} & \textbf{0.413} & \textbf{0.331} \\
\bottomrule
\end{tabular}
}
\caption{ChartMuseum CVA by reasoning type and model. CVA is conditioned on original-chart success. Reasoning-type names follow the ChartMuseum scheme.}
\label{tab:appendix-reasoning-types}
\end{table}

Table~\ref{tab:appendix-reasoning-types} reports the per-model counterpart to Figure~\ref{fig:chartmuseum-reasoning-types}, using the original ChartMuseum reasoning-type annotations.
It uses the same conditional framing as the main CVA result: variants are included only for families where the model answered the original chart correctly.
Thus, the comparison focuses on generalizability after original-chart success rather than static difficulty alone.

\subsection{ChartMuseum Failure Case Studies}
\label{sec:appendix-chartmuseum-case-studies}

We compare representative generated rationales for the first two ChartMuseum rows in Figure~\ref{fig:failure-case-studies}.
The first case illustrates stale prediction (SP) with a pattern consistent with original-chart information reuse and possible memorization.
The second illustrates noisy update (NU), where models move away from the original answer but still fail to ground the counterfactual variant correctly.

\begin{figure}[t]
\centering
\begin{minipage}{\columnwidth}
\textbf{Original Chart}
\begin{mytextbox}
\scriptsize\raggedright
\textbf{Gemini 2.5 Pro.}
The chart uses a color-coded legend.
``Time Wasters'' are represented by red.
The two red activities are Watching TV and Going to Concerts.
\textcolor{myblue}{Watching TV begins in early childhood, around age 6, and continues through adulthood past age 32.}
Going to Concerts starts much later, around age 18.
\textcolor{myblue}{Therefore, Watching TV spans the longest duration among Time Wasters.}
\textbf{Answer:} Watching TV.
\end{mytextbox}

\begin{mytextbox}
\scriptsize\raggedright
\textbf{GPT-5.4.}
The red bands represent the ``Time Wasters.''
Comparing the labeled red activities, \textcolor{myblue}{``Watching TV'' stretches across the widest age range, starting in early childhood and continuing into adulthood, longer than ``Video Games'' or ``Going to Concerts.''}
\textbf{Answer:} Watching TV.
\end{mytextbox}

\textbf{Counterfactual Variant}
\begin{mytextbox}
\scriptsize\raggedright
\textbf{Gemini 2.5 Pro.}
The legend indicates that ``Time Wasters'' are represented by red.
The red activities are Watching TV, Video Games, and Going to Concerts.
\textcolor{red}{Watching TV starts in early childhood, around age 4, and continues to the end of the chart at age 32.}
Video Games begins around age 7 and seems to end around age 21.
Going to Concerts starts in the late teens, around age 17, and continues until about age 31.
\textcolor{red}{Comparing start and end points, Watching TV has the longest duration.}
\textbf{Answer:} Watching TV.
\end{mytextbox}

\begin{mytextbox}
\scriptsize\raggedright
\textbf{GPT-5.4.}
The ``Time Wasters'' category is shown in red. Comparing the red-labeled activities, 
\textcolor{red}{``Watching TV'' stretches across the widest age range, starting in early childhood and continuing into early adulthood, longer than ``Video Games'' or ``Going to Concerts.''}
\textbf{Answer:} Watching TV.
\end{mytextbox}
\end{minipage}
\caption{Additional generated rationales for the first case study in \Cref{fig:failure-case-studies}. Highlighted spans show key reasoning points for the original chart (blue) and their incorrect reuse on the counterfactual variant (red).}
\label{fig:appendix-chartmuseum-case1-responses}
\end{figure}

\begin{figure}[!t]
\centering
\begin{minipage}{\columnwidth}
\textbf{Counterfactual Variant}
\begin{mytextbox}
\scriptsize\raggedright
\textbf{GPT-5.4.}
The metals price index is shown by the red line.
\textcolor{mygreen}{Scanning for its highest point, the peak occurs near the end of 2024, specifically in December 2024.}
\textbf{Answer:} December 2024.
\end{mytextbox}

\begin{mytextbox}
\scriptsize\raggedright
\textbf{Claude Sonnet 4.6.}
Looking at the red line (Metals), \textcolor{red}{the metals line appears to reach its highest point around November--December 2024}, where it climbs to approximately 113--114, the highest point the red line reaches throughout the chart period.
\textbf{Answer:} November 2024.
\end{mytextbox}

\begin{mytextbox}
\scriptsize\raggedright
\textbf{Gemini 2.5 Pro.}
The legend indicates that the red line represents Metals.
The question asks for the highest point of this red line.
By visually scanning the line, its highest value is reached late in 2024.
\textcolor{red}{Tracing this peak down to the x-axis, it aligns with the ``Nov'' tick mark, after ``Aug 2024'' and before ``Feb 2025.''}
\textbf{Answer:} Nov 2024.
\end{mytextbox}
\end{minipage}
\caption{Counterfactual variant responses for the metals price case. Highlighted spans mark the correct visual-to-text grounding (green) and the incorrect grounding to the nearby ``Nov'' tick (red).}
\label{fig:appendix-chartmuseum-case2-responses}
\end{figure}

\paragraph{Case 1: Stale prediction and possible memorization.}

In the first ChartMuseum case, the gold answer changes, but the representative model responses keep the original prediction.
The additional traces follow the same pattern as the Claude Sonnet 4.6 traces in Figure~\ref{fig:case1-claude-answer-comparison}: the counterfactual reasoning reuses duration claims that better describe the original chart.
For GPT-5.4, the reuse is nearly word-for-word: both traces say that ``Watching TV'' spans the widest age range, starts in early childhood, continues into adulthood, and is longer than ``Video Games'' or ``Going to Concerts.''
Gemini 2.5 Pro is less exact, but preserves the same original-chart explanation after the counterfactual variant changes the longest Time Waster.
This pattern is consistent with possible memorization or other reliance on parametric knowledge of the original chart or QA pair, while not ruling out a local visual error.

\paragraph{Case 2: Noisy update from incomplete visual-to-text grounding.}
In the second case, the gold answer changes from ``May 2024'' to ``Dec 2024,'' and \Cref{fig:appendix-chartmuseum-case2-responses} presents the generated rationales from each model.
GPT-5.4 updates correctly, while Claude Sonnet 4.6 and Gemini 2.5 Pro update to ``Nov 2024,'' producing noisy updates.
Both incorrect responses locate the late-2024 peak region but anchor it to the nearby visible ``Nov'' tick instead of inferring the missing December label.
Claude Sonnet 4.6 collapses a November--December localization to November, while Gemini 2.5 Pro explicitly traces the peak to the displayed ``Nov'' tick.

\newcommand{\charxivcasefigure}[9]{\begin{figure*}[t!]
\centering
\begin{minipage}[t]{0.48\textwidth}
    \centering
    \includegraphics[width=\textwidth,height=0.27\textheight,keepaspectratio]{#1}
\end{minipage}
\hfill
\vrule{}
\hfill
\begin{minipage}[t]{0.48\textwidth}
    \centering
    \includegraphics[width=\textwidth,height=0.27\textheight,keepaspectratio]{#2}
\end{minipage}
\begin{minipage}[t]{0.48\textwidth}
    \begin{flushleft}
        \scriptsize
        \textbf{Question:} #3\\
        \textbf{Original.} #4
    \end{flushleft}
\end{minipage}
\hfill
\vrule{}
\hfill
\begin{minipage}[t]{0.48\textwidth}
    \begin{flushleft}
        \scriptsize
        \textbf{Question:} #3\\
        \textbf{Counterfactual.} #5
    \end{flushleft}
\end{minipage}
\begin{flushleft}
    \scriptsize
    \textbf{Update outcome rates:} #6
\end{flushleft}
\begin{flushleft}
    \scriptsize
    \textbf{Case note:} #7
\end{flushleft}
\captionsetup{skip=3pt}
\caption{#8}
\label{#9}
\end{figure*}
}

\newcommand{\charxivcompactcase}[9]{
\begin{minipage}[t]{0.48\textwidth}
    \centering
    \includegraphics[width=\textwidth,height=0.28\textheight,keepaspectratio]{#1}
\end{minipage}
\hfill
\vrule{}
\hfill
\begin{minipage}[t]{0.48\textwidth}
    \centering
    \includegraphics[width=\textwidth,height=0.28\textheight,keepaspectratio]{#2}
\end{minipage}
\begin{minipage}[t]{0.48\textwidth}
    \scriptsize\raggedright
    \textbf{Question:} #3\\
    \textbf{Original.} #4
\end{minipage}
\hfill
\vrule{}
\hfill
\begin{minipage}[t]{0.48\textwidth}
    \scriptsize\raggedright
    \textbf{Question:} #3\\
    \textbf{Counterfactual.} #5
\end{minipage}
\par\smallskip
{\scriptsize\textbf{Update outcome rates:} #6\par}
{\scriptsize\textbf{Case note:} #7\par}
\caption{#8}
\label{#9}
\par\vspace{5mm}
}

\subsection{CharXiv Failure Case Studies}
\label{sec:appendix-charxiv-case-studies}

CharXiv does not provide reasoning-type annotations, so we group selected NU/SP cases using ChartMuseum-inspired visual task categories~\cite{tang2026chartmuseumtestingvisualreasoning}, with more specific labels when needed.
We surface candidates using original-to-variant accuracy gaps and NU/SP rates, then manually retain interpretable failures that provide category coverage.
For each family, the dominant wrong-update mode is computed over proprietary model--seed pairs after original-chart success.
Table~\ref{tab:appendix-charxiv-case-study-statistics} reports the resulting counts, which characterize this diagnostic set rather than benchmark-wide prevalence.
Figures~\ref{fig:appendix-charxiv-case-study-trajectory-risk}--\ref{fig:appendix-charxiv-case-study-threshold-scatter} show representative pairs with answers, case notes, and correctness colors.
\makeatletter
\setlength{\@dblfptop}{0pt}
\makeatother

\begin{table}[!ht]
\centering
\footnotesize
\setlength{\tabcolsep}{4pt}
\resizebox{\columnwidth}{!}{\begin{tabular}{@{}lrrr@{}}
\toprule
Category & NU & SP & Total \\
\midrule
Trajectory tracking and judgment & 11 & 1 & 12 \\
Value identification or thresholding & 6 & 0 & 6 \\
Spatial visual comparison & 5 & 0 & 5 \\
Symbol-guided visual comparison & 3 & 1 & 4 \\
Other visually grounded cases & 1 & 1 & 2 \\
\midrule
\textbf{Total} & \textbf{26} & \textbf{3} & \textbf{29} \\
\bottomrule
\end{tabular}
}
\caption{CharXiv NU/SP case-study summary. Rows count selected families by visual task category. NU and SP indicate the dominant wrong-update mode over proprietary model-seed pairs after original-chart success.}
\label{tab:appendix-charxiv-case-study-statistics}
\end{table}

\begin{figure*}[p]
\centering
\charxivcompactcase
{}
{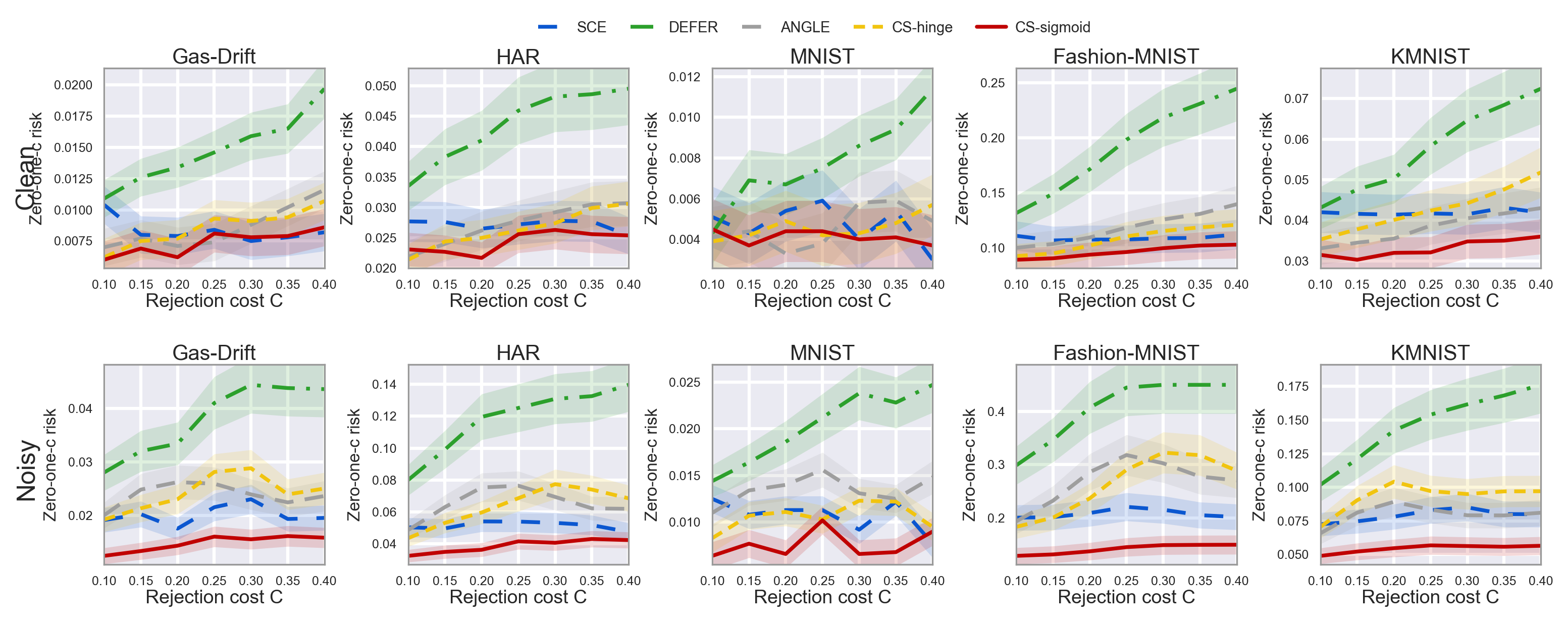}
{In the Noisy MNIST condition, which model has the greatest drop in zero-one risk at rejection cost $c=0.35$?}
{Gold: CS-hinge.\\ GPT-5.4: CS-hinge. Claude Sonnet 4.6: CS-hinge. Gemini 2.5 Pro: CS-hinge.}
{Gold: DEFER.\\ GPT-5.4: \textcolor{red}{ANGLE}. Claude Sonnet 4.6: \textcolor{red}{CS-sigmoid}. Gemini 2.5 Pro: \textcolor{red}{CS-sigmoid}.}
{CU 0.33 / NU 0.62 / SP 0.05.}
{Trajectory tracking and judgment. The answer requires first locating the Noisy MNIST panel among several datasets, then reading several model curves at the same rejection cost. The target is not the lowest curve overall, but the largest local decrease around $c=0.35$, so small mistakes in curve identity or local slope can change the answer.}
{CharXiv case study: trajectory tracking in risk curves.}
{fig:appendix-charxiv-case-study-trajectory-risk}

\charxivcompactcase
{}
{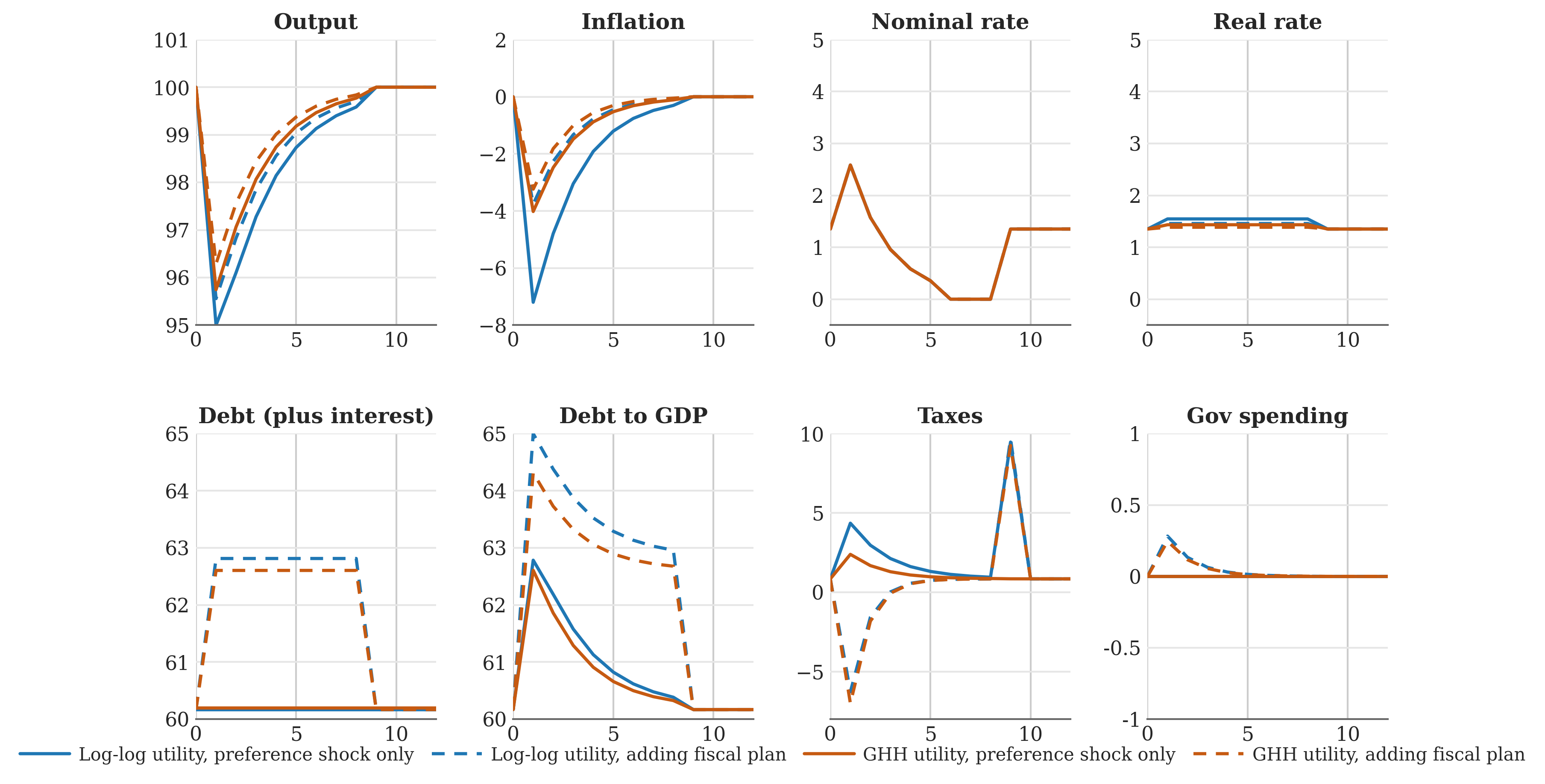}
{Between $x=0$ and $x=5$, which subplot has curves with the most negative relation to the inflation subplot?}
{Gold: Real rate.\\ GPT-5.4: Real rate. Gemini 2.5 Pro: Real rate. GPT-5.4 mini: Real rate.}
{Gold: Nominal rate.\\ GPT-5.4: \textcolor{red}{Taxes}. Gemini 2.5 Pro: \textcolor{red}{Debt to GDP}. GPT-5.4 mini: \textcolor{red}{Output}.}
{CU 0.47 / NU 0.53 / SP 0.00.}
{Trajectory tracking and judgment. The question constrains the comparison to the interval $x=0$ to $x=5$, so the model must compare local relationships rather than the full-subplot trend. It also requires relating each candidate subplot to the inflation subplot, making the counterfactual sensitive to panel binding and local trend comparison errors.}
{CharXiv case study: trajectory comparison across economic subplots.}
{fig:appendix-charxiv-case-study-trajectory-subplots}

\charxivcompactcase
{}
{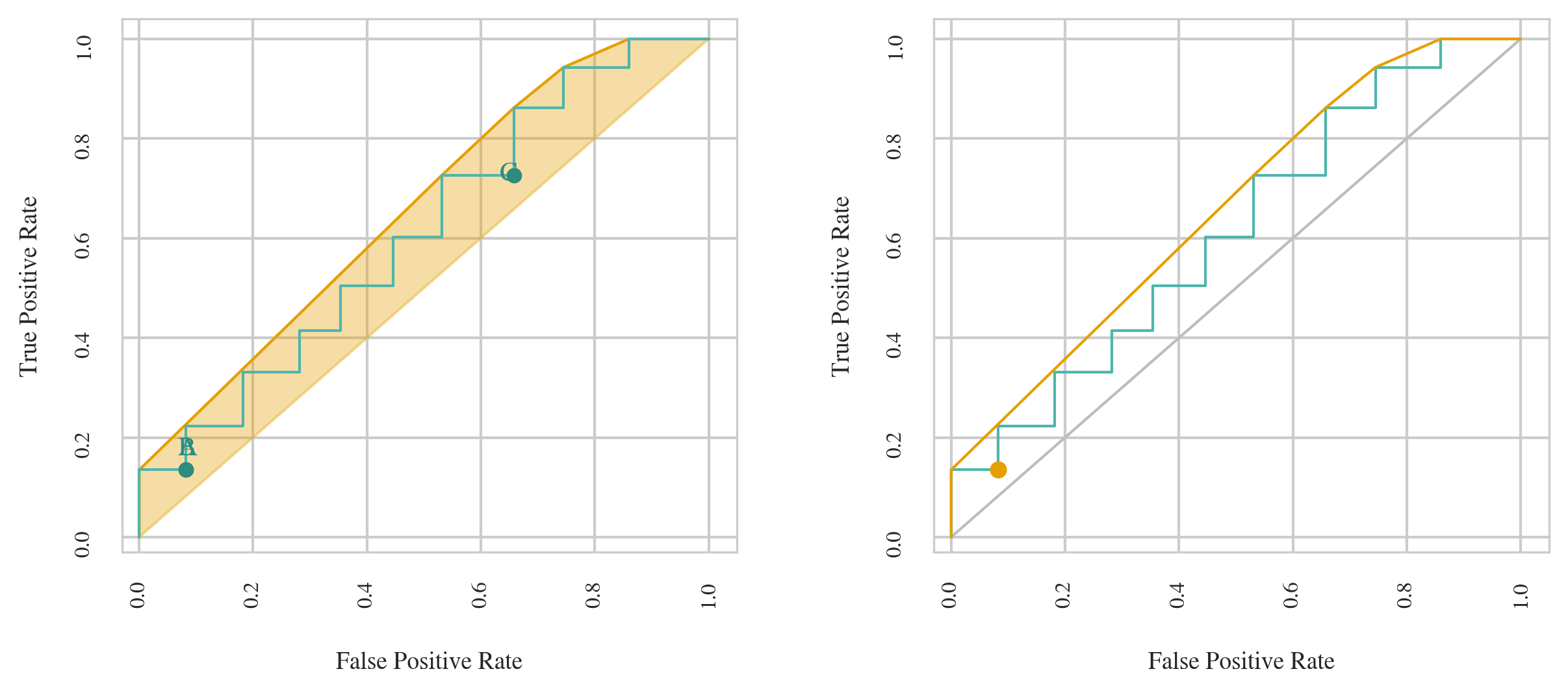}
{How many steps separate point B and C in the figure on the left?}
{Gold: 4.\\ GPT-5.4: 4. Gemini 2.5 Pro: 4. GPT-5.4 mini: 4.}
{Gold: 6.\\ GPT-5.4: \textcolor{red}{10}. Gemini 2.5 Pro: \textcolor{red}{4}. GPT-5.4 mini: \textcolor{red}{5}.}
{CU 0.27 / NU 0.70 / SP 0.03.}
{Spatial visual comparison. The answer is a count over the discrete staircase path, not a continuous distance between points B and C. The model must identify the correct panel, follow the step sequence in order, and avoid substituting nearby axis values or the visually salient diagonal baseline for the path length.}
{CharXiv case study: spatial step counting.}
{fig:appendix-charxiv-case-study-spatial-steps}

\end{figure*}

\begin{figure*}[p]
\centering

\charxivcompactcase
{}
{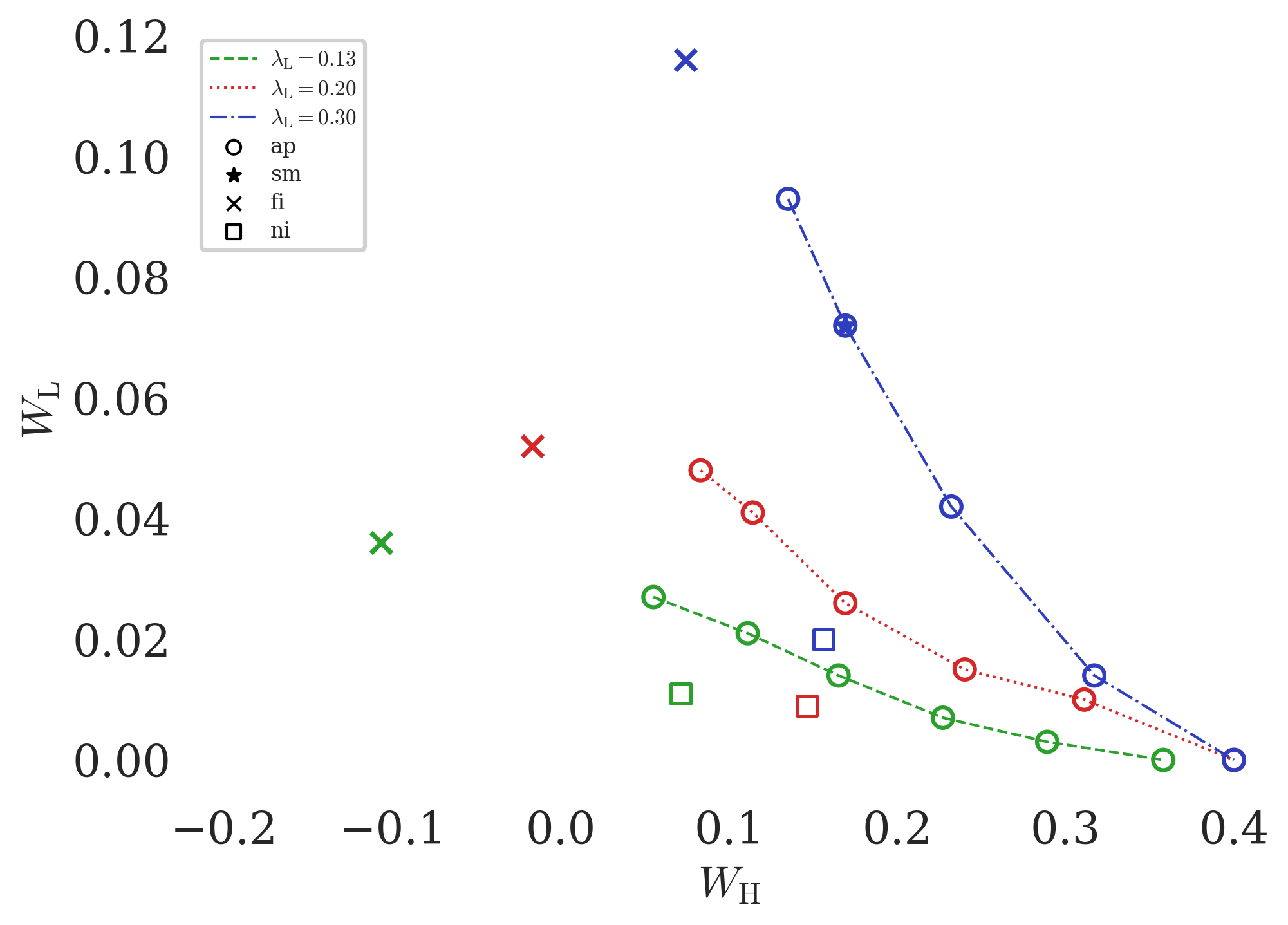}
{Which line is furthest away from its $f_i$ value from the $W_H$ axis?}
{Gold: $\lambda_L=0.13$.\\ GPT-5.4: $\lambda_L=0.13$. Claude Sonnet 4.6: $\lambda_L=0.13$. Gemini 2.5 Flash: $\lambda_L=0.13$.}
{Gold: $\lambda_L=0.13$.\\ GPT-5.4: \textcolor{red}{$\lambda_L=0.30$}. Claude Sonnet 4.6: \textcolor{red}{$\lambda_L=0.30$}. Gemini 2.5 Flash: \textcolor{red}{$\lambda_L=0.30$}.}
{CU 0.30 / NU 0.70 / SP 0.00.}
{Spatial visual comparison. The answer depends on binding each colored trajectory to its corresponding $f_i$ marker and comparing the marker's horizontal distance from the $W_H$ axis. Because the curves, open markers, and $f_i$ symbols are close together, a model can correctly recognize the chart type while still comparing the wrong visual object.}
{CharXiv case study: spatial line-marker binding.}
{fig:appendix-charxiv-case-study-spatial-lines}

\charxivcompactcase
{}
{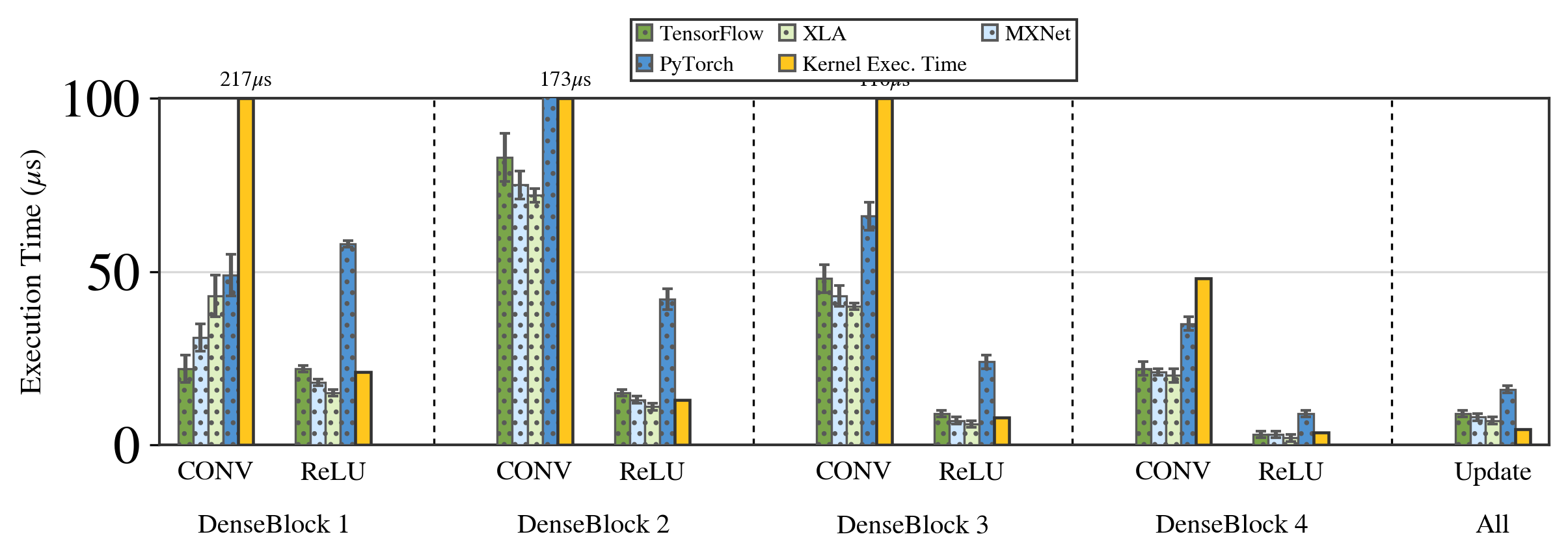}
{Which library has the highest reduction in CONV execution time from DenseBlock 1 to DenseBlock 4?}
{Gold: PyTorch.\\ GPT-5.4: PyTorch. Claude Sonnet 4.6: PyTorch. Gemini 2.5 Pro: PyTorch.}
{Gold: XLA.\\ GPT-5.4: \textcolor{red}{MXNet}. Claude Sonnet 4.6: \textcolor{red}{PyTorch}. Gemini 2.5 Pro: \textcolor{red}{MXNet}.}
{CU 0.34 / NU 0.60 / SP 0.06.}
{Symbol-guided visual comparison. The question asks for a reduction within one operation type, so the model must preserve the legend identity for each library and compare only the CONV bars from DenseBlock 1 to DenseBlock 4. ReLU bars, update bars, and the repeated library colors provide plausible distractors.}
{CharXiv case study: legend-bound bar comparison.}
{fig:appendix-charxiv-case-study-symbol-bars}

\charxivcompactcase
{}
{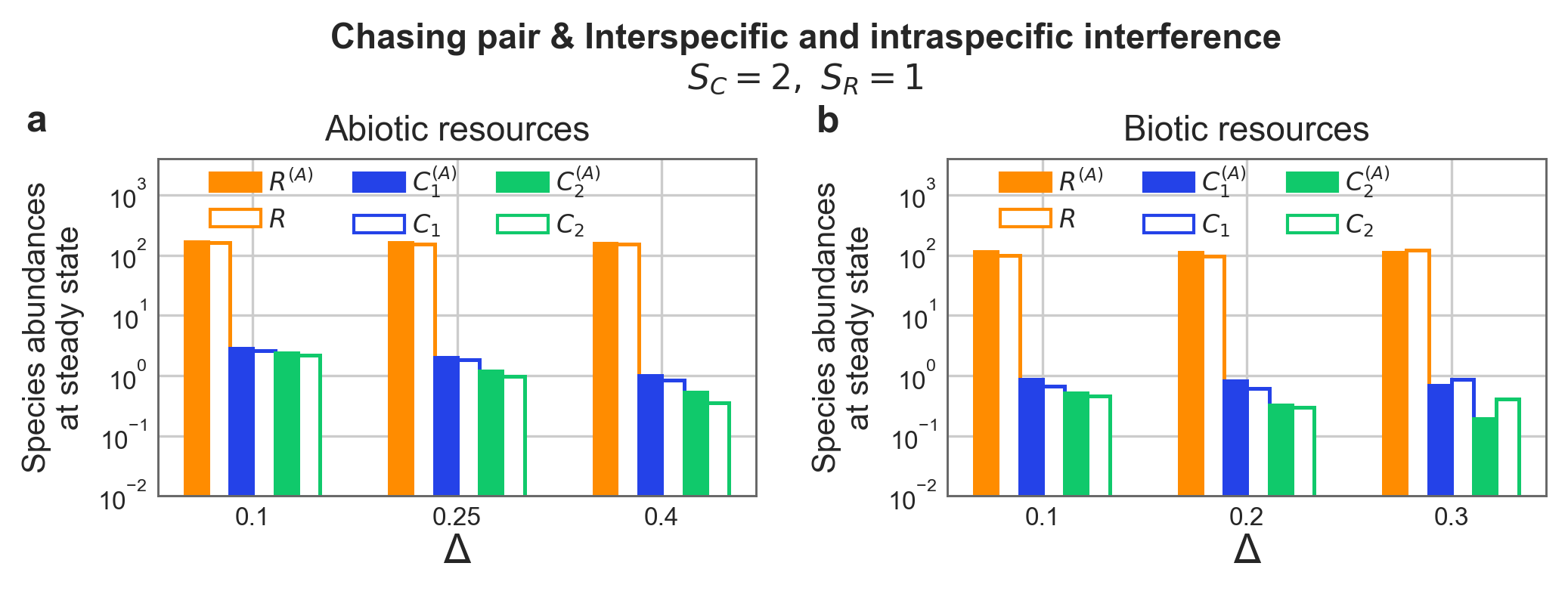}
{Which subplot has each non-(A) data type always smaller than its (A) counterpart?}
{Gold: Biotic resources.\\ GPT-5.4: Biotic resources. Claude Sonnet 4.6: Biotic resources. Gemini 2.5 Pro: Biotic resources.}
{Gold: Abiotic resources.\\ GPT-5.4: \textcolor{green}{Abiotic resources}. Claude Sonnet 4.6: \textcolor{red}{Biotic resources}. Gemini 2.5 Pro: \textcolor{green}{Abiotic resources}.}
{CU 0.38 / NU 0.14 / SP 0.48.}
{Symbol-guided visual comparison. The answer is a universal comparison within a subplot: every non-(A) bar must be smaller than its matching (A) counterpart. This requires matching paired encodings across several categories, so a single violated pair should rule out a subplot even if its overall pattern looks similar.}
{CharXiv case study: paired bar matching.}
{fig:appendix-charxiv-case-study-symbol-pairs}

\end{figure*}

\begin{figure*}[p]
\centering

\charxivcompactcase
{}
{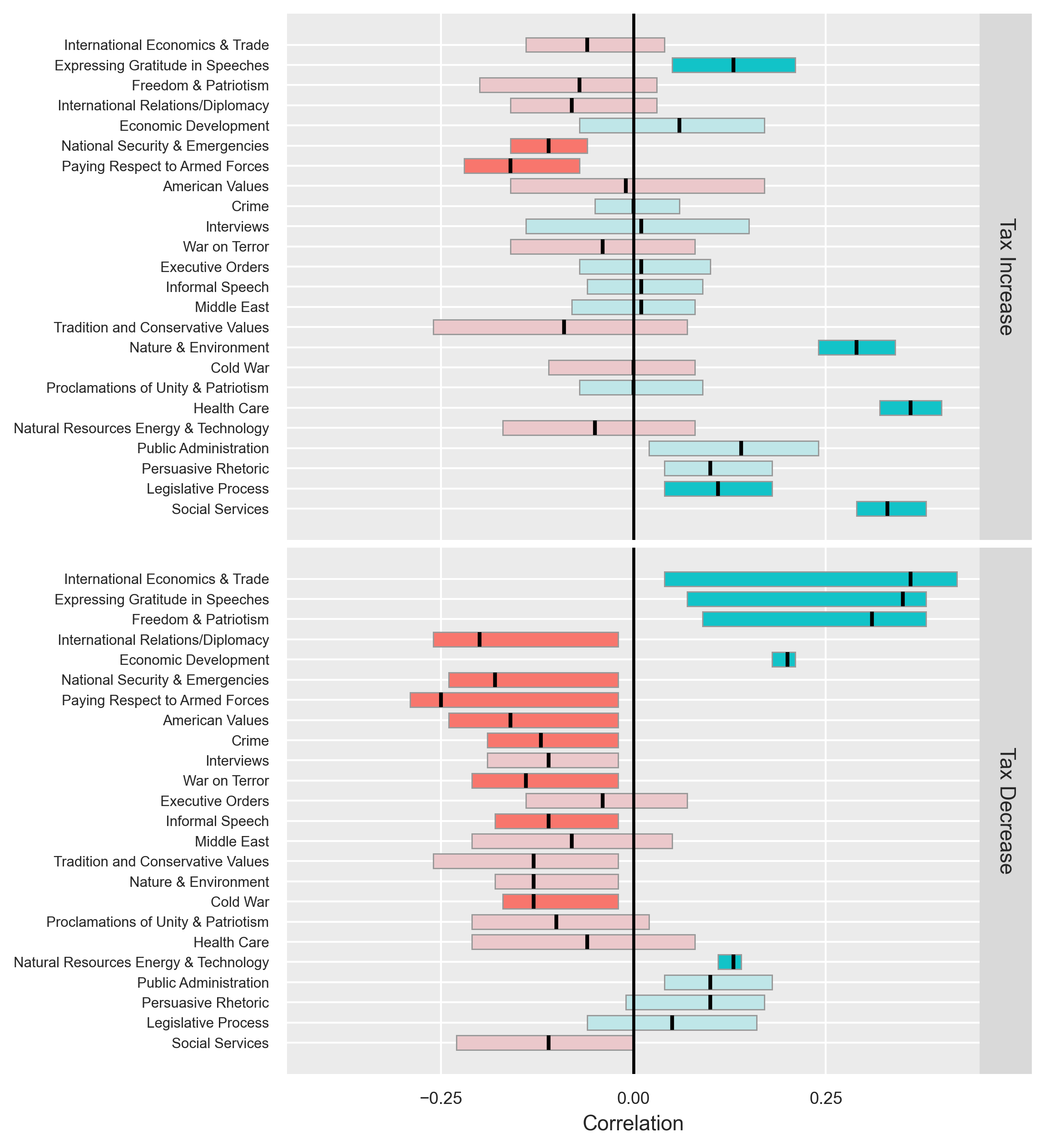}
{How many bars exceed 0.25 in the Tax Decrease chart?}
{Gold: 2.\\ GPT-5.4: 2. Gemini 2.5 Pro: 2. GPT-4o: 2.}
{Gold: 3.\\ GPT-5.4: \textcolor{red}{4}. Gemini 2.5 Pro: \textcolor{red}{5}. GPT-4o: \textcolor{red}{1}.}
{CU 0.45 / NU 0.50 / SP 0.05.}
{Value identification and thresholding. The question requires selecting the Tax Decrease facet, applying the 0.25 cutoff, and counting only bars that exceed that threshold. Nearby panels share the same visual scale, and several bars sit close to the cutoff, so both panel-binding and thresholding errors can change the count.}
{CharXiv case study: thresholded bar counting.}
{fig:appendix-charxiv-case-study-threshold-bars}

\charxivcompactcase
{}
{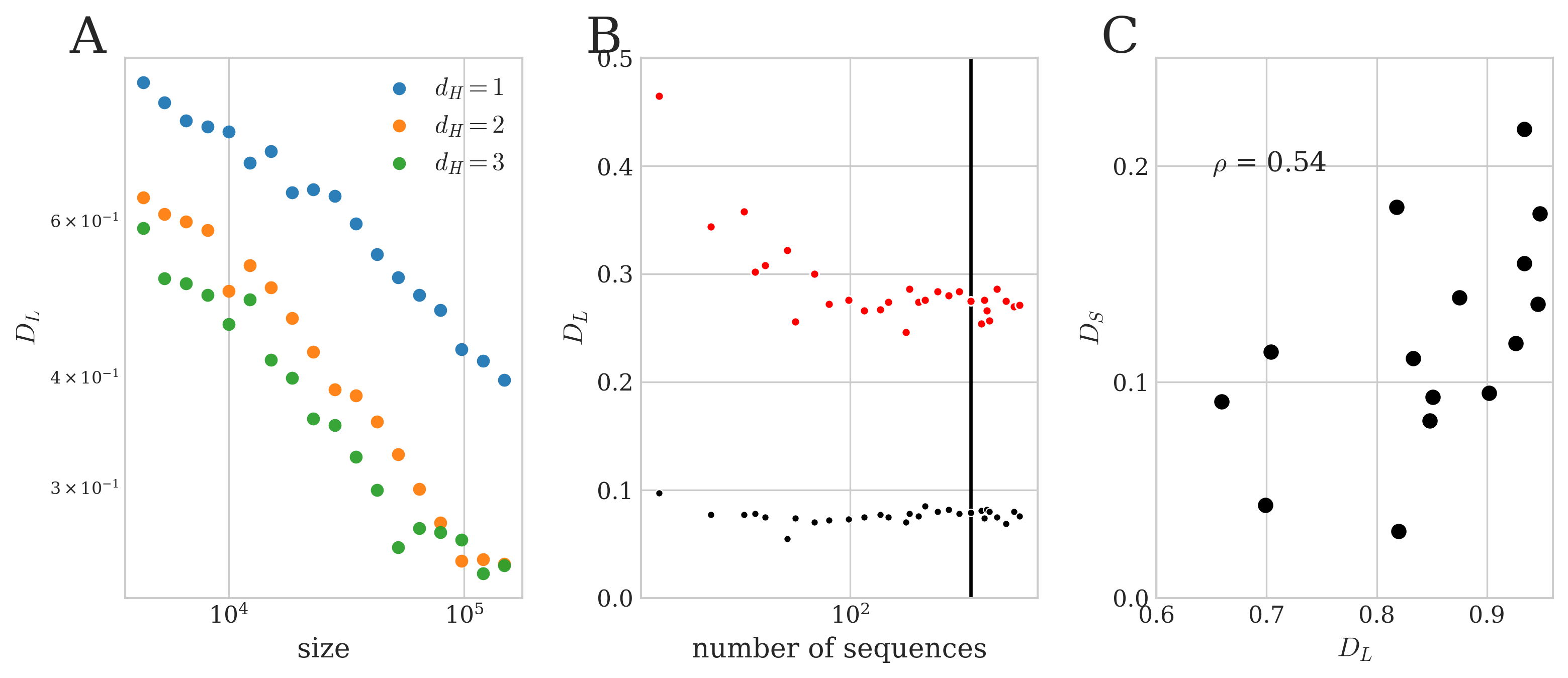}
{How many data points have a $D_{JS}$ value less than 0.1?}
{Gold: 3.\\ GPT-5.4: 3. GPT-5.4 mini: 3.}
{Gold: 6.\\ GPT-5.4: \textcolor{green}{6}. GPT-5.4 mini: \textcolor{red}{5}.}
{CU 0.45 / NU 0.45 / SP 0.10.}
{Value identification and thresholding. The model must identify the panel containing the $D_{JS}$ values, apply a strict threshold at 0.1, and count individual points rather than infer from the overall trend or correlation annotation. Points near the cutoff make small value-reading errors sufficient to change the answer.}
{CharXiv case study: thresholded scatter-point counting.}
{fig:appendix-charxiv-case-study-threshold-scatter}
\end{figure*}